\documentclass[lettersize,journal]{IEEEtran}
\usepackage{amsmath,amsfonts}
\usepackage{algorithmic}
\usepackage{algorithm}
\usepackage{array}
\usepackage{array} 
\newcolumntype{M}[1]{>{\centering\arraybackslash}m{#1}}
\usepackage{textcomp}
\usepackage{stfloats}
\usepackage{url}
\usepackage[caption=false]{subfig}
\usepackage{verbatim}
\usepackage{subcaption}
\usepackage{booktabs}
\usepackage{graphicx}
\usepackage{cite}
\usepackage{xcolor}
\title{Atomic Depth Estimation From Noisy Electron Microscopy Data Via Deep Learning 
}
\author{%
  Matan Leibovich\IEEEauthorrefmark{1},
  Mai Tan\IEEEauthorrefmark{2},
Ramon Manzorro\IEEEauthorrefmark{2},
  Adria Marcos Morales\IEEEauthorrefmark{3},
  Sreyas Mohan\IEEEauthorrefmark{3},
  Peter A. Crozier\IEEEauthorrefmark{2},
  Carlos Fernandez\textendash Granda\IEEEauthorrefmark{1}\IEEEauthorrefmark{3}%
\thanks{\IEEEauthorrefmark{1}Courant Institute of Mathematical Sciences, New York University, New York, NY, USA.}%
\thanks{\IEEEauthorrefmark{2}School for Engineering of Matter, Transport, and Energy, Arizona State University, Tempe, AZ, USA.}%
\thanks{\IEEEauthorrefmark{3}Center for Data Science, New York University, New York, NY, USA.}%
}
\begin{document}

\maketitle
    \begin{abstract}
We present a novel approach for extracting 3D atomic-level information from transmission electron microscopy (TEM) images affected by significant noise. The approach is based on formulating depth estimation as a semantic segmentation problem. We address the resulting segmentation problem by training a deep convolutional neural network to generate pixel-wise depth segmentation maps using simulated data corrupted by synthetic noise. The proposed  method was applied to estimate the depth of atomic columns in CeO2 nanoparticles from simulated images and real-world TEM data. Our experiments show that the resulting depth estimates are accurate, calibrated and robust to noise. %In addition, we show that the deep neural network relies on long-range information to accurately predict the local 3D structure. % , due to the nonlinear dependence of the image intensity on the column depth. %Our findings assert that the network exhibits advantageous features, such as proper calibration and robustness against the signal to noise ratio (SNR) it was trained on. We demonstrate the application of the network for exploration of new electrochemical processes.
\end{abstract}

    \section{Introduction}
    \label{sec:introduction}
    
    %We are interested in estimating the three dimensional structure of nanoparticles using transmission electron microscopy images. 
    Estimating the three-dimensional (3D) structure of objects in microscopic imaging is important for a wide-ranging array of applications. Characterizing the 3D structure of molecules is essential for the discovery of new treatments in medicine and epidemiology \cite{punjani2016building}. Determining the atomic structure of nanoparticles can provide information about their functionality, which is key to design materials with novel properties \cite{kalinin2015big}. The problem is inherently complex, since imaging data is two-dimensional and often acquired under constraints that limit its signal-to-noise ratio.
    
    In this work, we consider the problem of extracting 3D information from images obtained by transmission electron microscopy (TEM). TEM is a powerful and versatile characterization technique that is used to probe the atomic-level structure and composition of a wide range of materials, such as catalysts and semiconductors~\cite{Smith2015,Crozier-insitu2016}. Traditional estimation of 3D structures, via electron tomography, involves tilting the sample into many different orientations (typically 20 or more), recording images at each orientation, and finally reconstructing a 3D image. This approach works well for static objects, but cannot be applied to dynamic objects experiencing changes on the timescale required to record the tilt series. This situation is prevalent in material sciences when studying the functionality of materials associated with structural dynamics that occur under working conditions. Examples include the surface of a catalyst, which may change when exposed to reactants, or phase changes that occur in battery or fuel cell materials when an electrical bias is applied. 
    
    In situ electron microscopy makes it possible to probe dynamic changes in the presence of a variety of applied stimuli \cite{de2011electron,mecklenburg2015nanoscale,hansen2016controlled,tao2016atomic}. To observe spatio-temporal dynamics, phase contrast imaging \cite{spence2013high}, coupled with direct electron detectors \cite{mcmullan2009detective,clough2016direct,maclaren2020detectors,plotkin2020hybrid,levin2021direct}, now allows atomic resolution images to be recorded at more than 1000 frames per second with sub-Angstrom spatial resolution. We have recently used this approach to investigate the atomic-level changes taking place on the surface of platinum nanoparticles~\cite{crozier2025visualizing}.  

%Combining high frame rates to observe structural dynamics with 3D electron tomography is very challenging. 

\begin{figure*}[htbp] 
\begin{tabular}{c c c c }
        \includegraphics[width=0.23\linewidth]{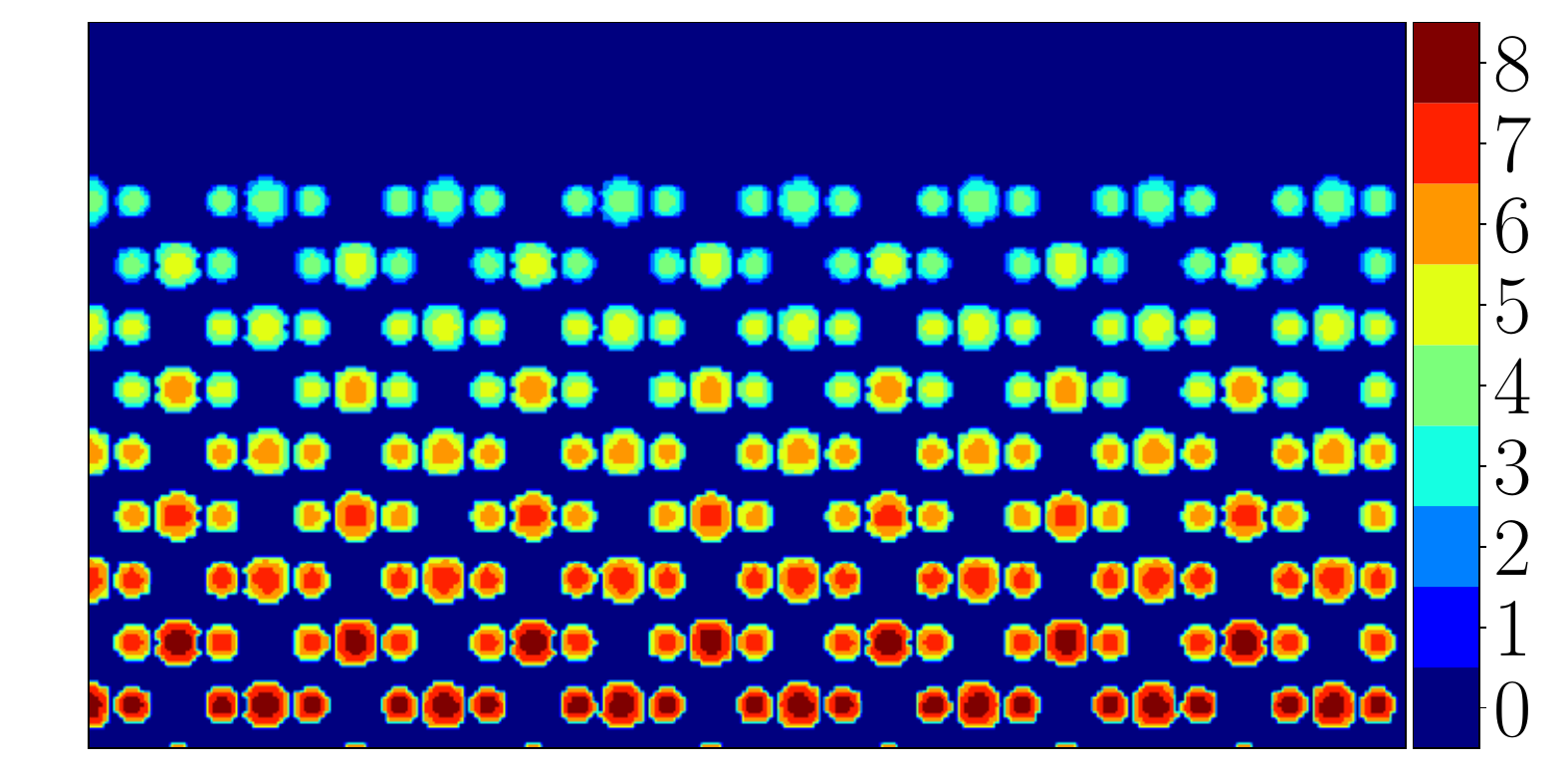}& \includegraphics[width=0.23\linewidth]{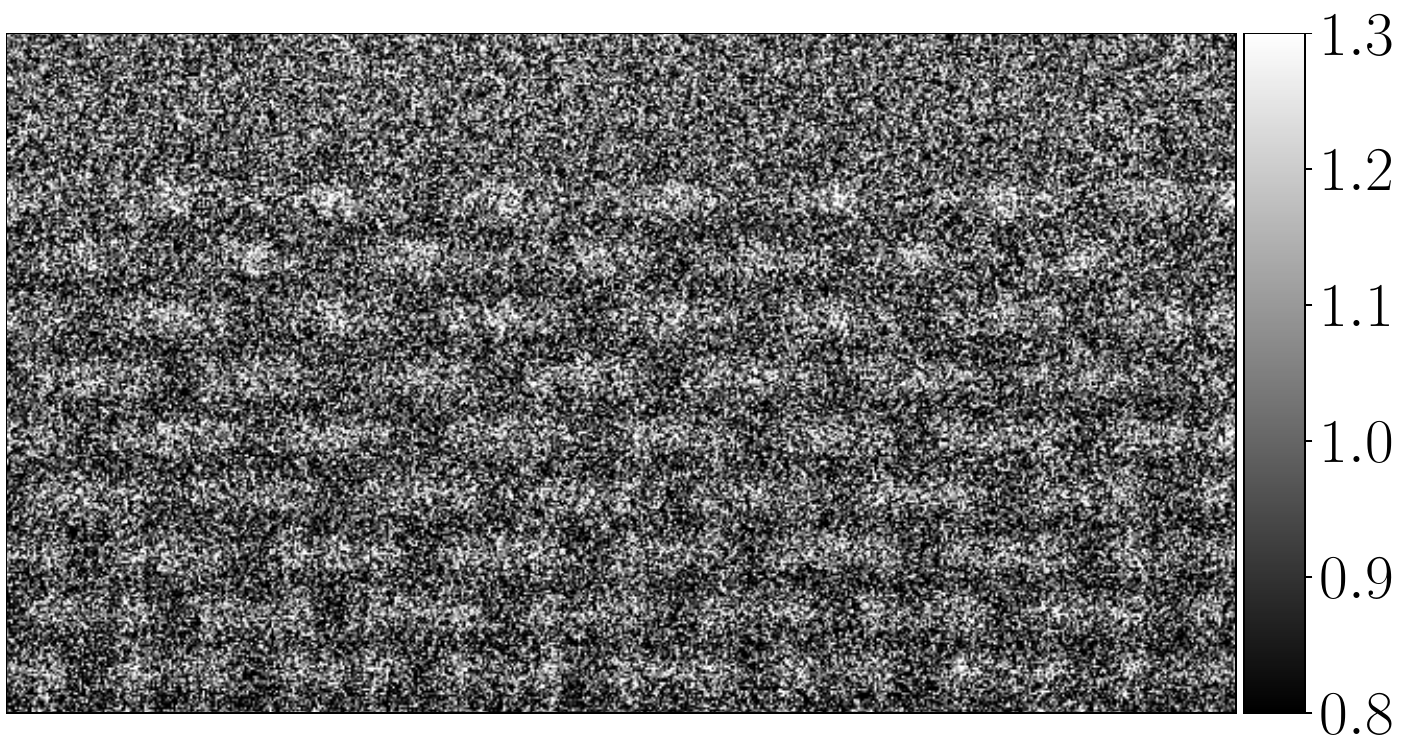}&
        \includegraphics[width=0.23\linewidth]{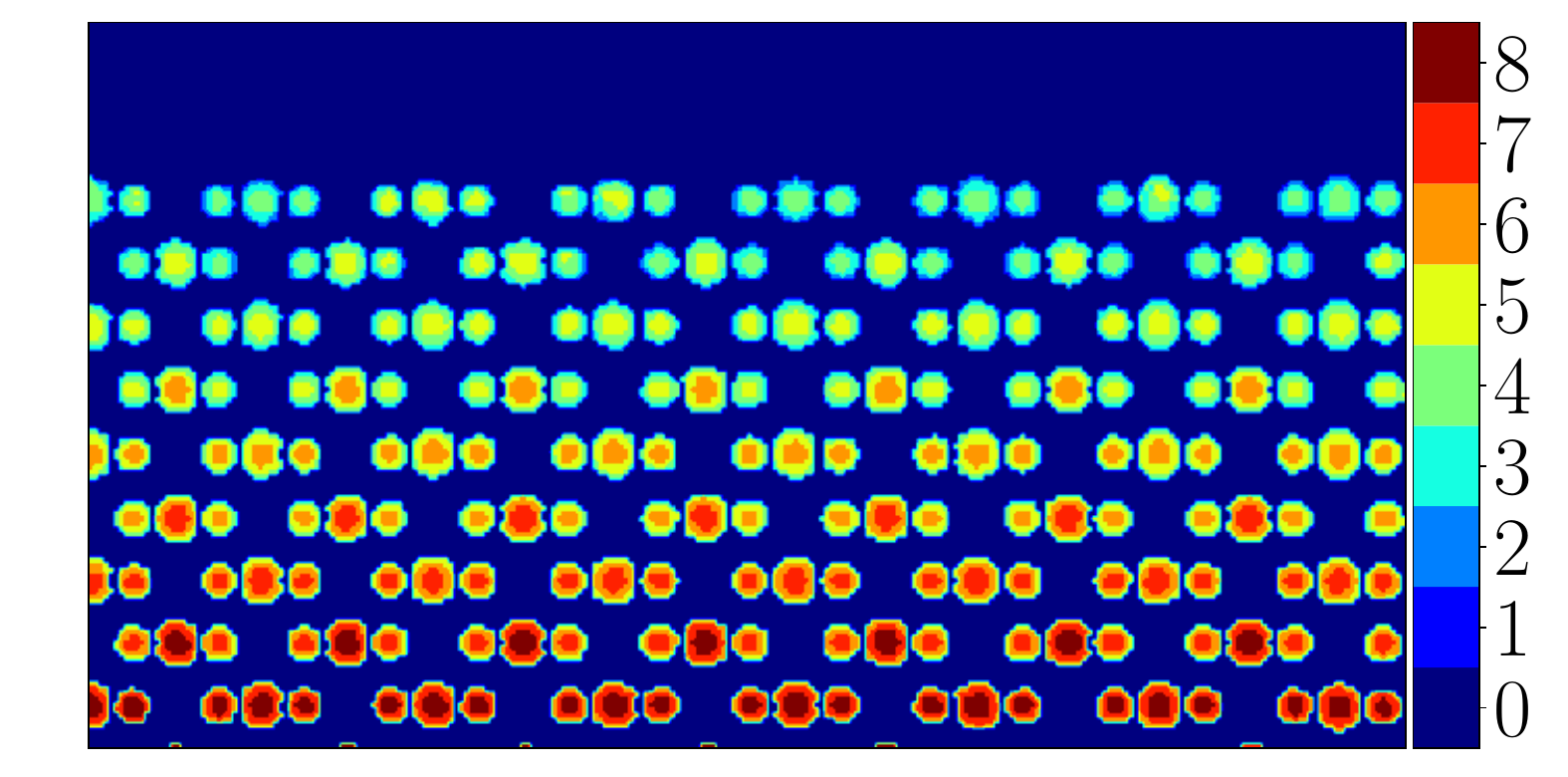}&
        \includegraphics[width=0.23\linewidth]{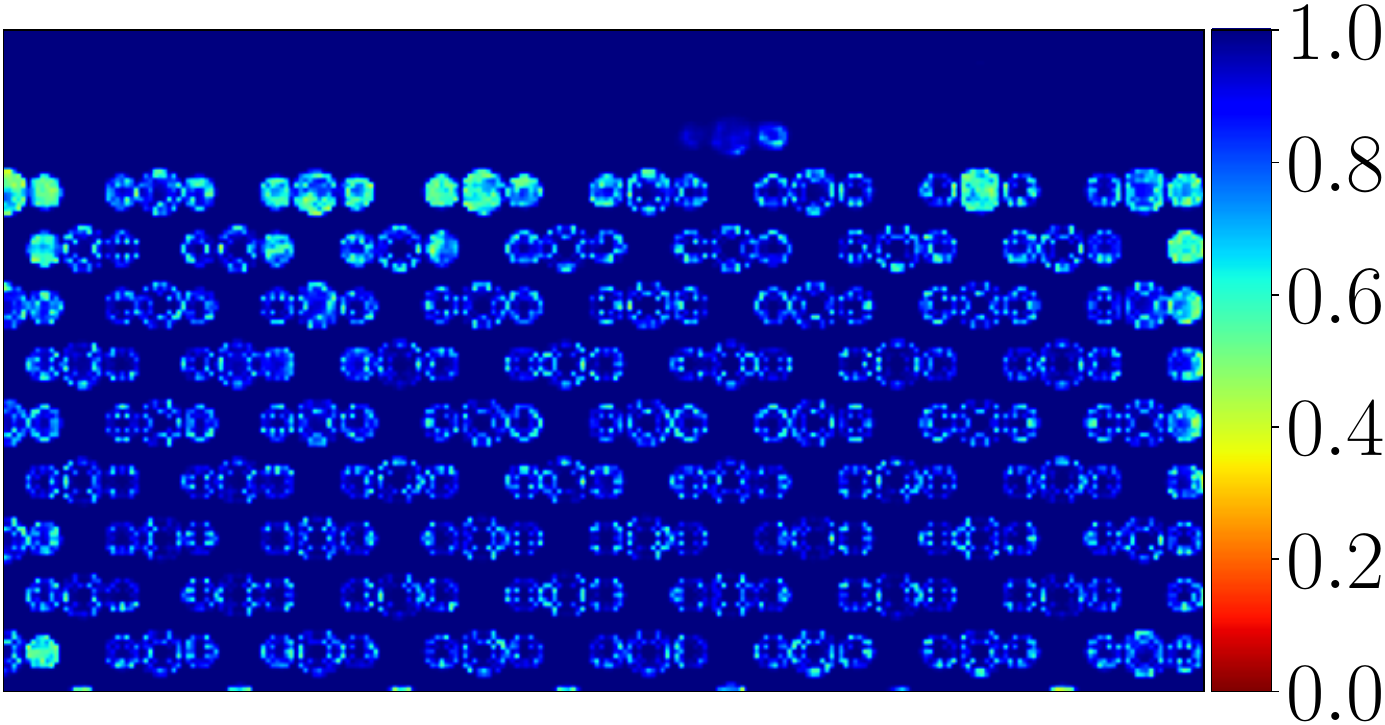}\\
        \small{$(a)$ Atomic-column depth}&\small{$(b)$ Noisy data} &
        \small{$(c)$ Estimate}&\small{$(d)$ Confidence }
\end{tabular}
\caption{\textbf{Depth estimation via segmentation.} 
%\textcolor{black}{(a) Simulated TEM image corresponding to a CeO2 nanoparticle in (110) zone axis orientation}. In this orientation the Ce and O atomic columns are distinct. 
(a) Atomic-column depth of a CeO2 nanoparticle in (110) zone axis orientation. In this orientation the Ce and O atomic columns are distinct. The thickness of the material increases from top to bottom. (b) Noisy data generated by corrupting the TEM image corresponding to (a) with Poisson noise to simulate real acquisition data. The Ce columns in the thinner columns at the top are bright and have fainter oxygen columns to the left and right. In the thicker columns at the bottom the contrast is reversed. (c) Estimate of the atomic-column depth obtained by the proposed deep-learning based method, where a neural network is trained to approximate the depth profile (a) from the noisy data (c). (d) Confidence score of the proposed model at each pixel, computed as defined in \eqref{eq:confidence}, which quantifies the uncertainty in the model estimates. The confidence is generally high, except on the boundary of the nanoparticle, where estimation is more challenging.}
\label{fig:example}
\end{figure*}

Motivated by the study of structural dynamics in materials, we propose \emph{SegDepth}, a framework to estimate 3D information from individual 2D images utilizing a deep learning segmentation model. 
%introduce \textbf{3D-EMnet}, a framework for extracting 3D information from atomic resolution phase contrast TEM images using semantic segmentation convolutional neural networks. Deep neural network have been successfully applied to problems involving processing and analyzing high dimensional data.  Machine learning based algorithms were used in processing images of different sources, such as medical images, and images encountered in biology, chemistry, physics, and materials science.
%Under this initial premise, 
Our goal is to use TEM images to estimate the depth of the atomic columns in nanoparticles. An atomic column is a linear arrangement of atoms stacked perpendicular to the imaging plane. Estimating how many atoms are in each column is an inverse problem, which requires relating the acquired electron image intensity pattern to the underlying atomic structure. The problem is complicated by several factors, illustrated in Figure~\ref{fig:example}. First, the non-linear nature of dynamical electron scattering, underlying the phase contrast image formation processes, makes the relation between intensity and the number of atoms non-linear over realistic sample thicknesses (0 – 5 nm) . Second, the signal-to-noise ratio (SNR) in the acquired time series images is extremely low, a consequence of the short exposure times required to achieve high temporal resolution and the limited electron dose rate, necessary to prevent radiation from damaging the material. 
\addtolength{\tabcolsep}{0pt} 
\addtolength{\tabcolsep}{5pt} 
    As a proof of concept for the proposed method, we focus on nanoparticles of CeO2, a reducible oxide system with applications in catalysis, fuel cells and memristors. For such applications, there is considerable interest in understanding oxygen transport and exchange, which may manifest itself as a change in the occupancy of the oxygen atomic column in time~\cite{trovarelli2017ceria}. This requires the ability to determine the oxygen column occupancy from noisy time-series data. 
    
    In order to train and evaluate the depth-estimation model, we designed a dataset of simulated TEM images of nanoparticle surfaces of varying thickness in high-symmetry zone axis orientation with respect to the incident electron beam. When the crystal is in a zone-axis orientation, peaks appear in the TEM image, representing columns of atoms in the structure. The intensity of the column is associated non-linearly with the number of atoms in the column and the depth of the column. The dataset was corrupted by Poisson noise to simulate real acquisition conditions. 
    
\begin{figure*}
    \centering
  \subfloat{%
       \includegraphics[width=0.69\linewidth]{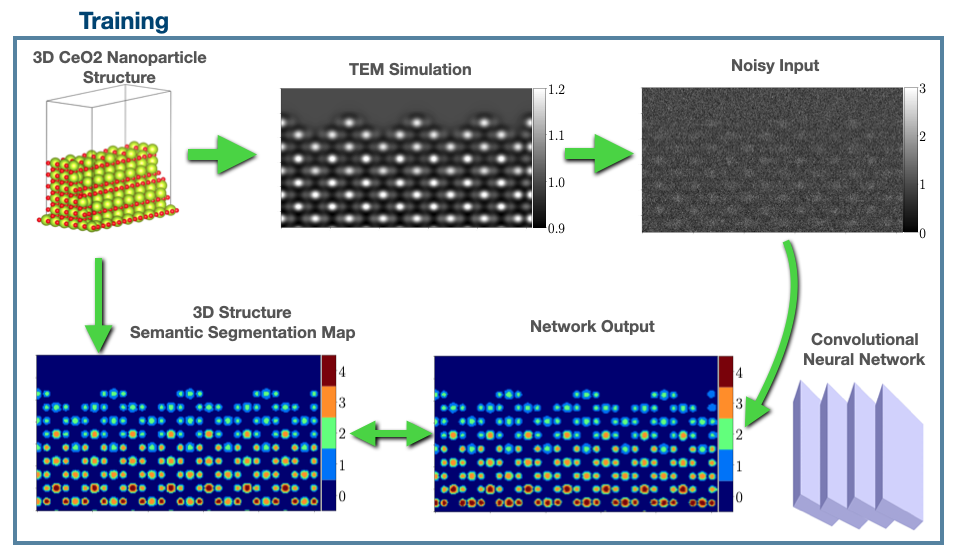}}\\\
    \subfloat{\includegraphics[width=0.69\linewidth]{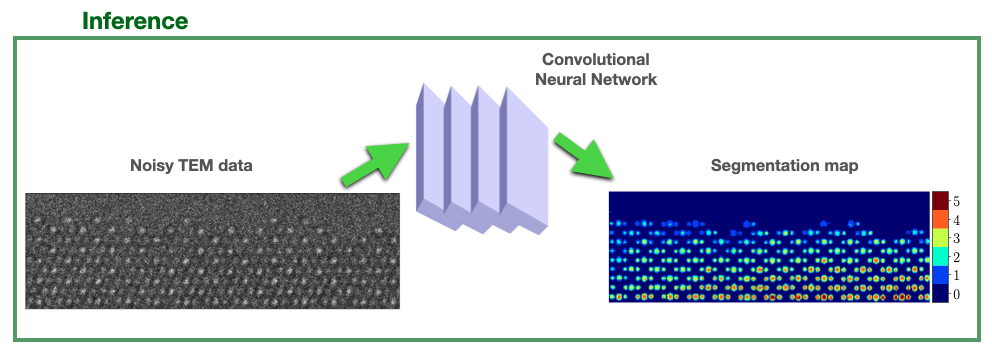}}
\caption{\textbf{The SegDepth framework.} \emph{Top:} A deep convolutional neural network is trained to estimate atomic-column depth using simulated data. Different 3D CeO2 nanoparticle structures are used to generate simulated TEM data, which are then corrupted with Poisson noise to produce the noisy input to the network. The 3D structure is also used to create segmentation maps that encode the atomic-column depth. The network is trained to approximate these segmentation maps. \emph{Bottom:} At inference, noisy TEM data is fed into the neural network to produce a segmentation map that estimates the atomic-column depth at every pixel.
} 
        \label{fig:scheme}
\end{figure*}

    As illustrated in Figure~\ref{fig:example}, we cast depth estimation as a segmentation problem, where the goal is to estimate the number of atoms or depth of the atomic column associated with each pixel in a noisy input image. Segmentation maps are generated using simulated data, where the column depth is known. These masks are used to train a deep convolutional neural network, which estimates the depth from corresponding simulated noisy TEM images. The network can then be applied to real data in order to investigate the dynamics of the atomic structure of nanoparticles as they interact with the ambient environment. This framework, which we call SegDepth, is illustrated in Figure~\ref{fig:scheme}. 

% This network is an important step forward in analyzing complex structures and one that is strictly facilitated by deep learning, as there are no other equivalent inverse solvers.

 \subsubsection*{Paper outline}
    %The paper is structured as follows:
    In \textbf{Section~\ref{sec:related_work}} we discuss related work. In \textbf{Section~\ref{sec:methodology}} we describe the proposed framework. In \textbf{Section~\ref{sec:simulated}} we provide a comprehensive evaluation of the framework using simulated data. In \textbf{Section~\ref{sec:real_dataset}} we report results on real data. 
    %In \textbf{Section~\ref{sec:analysis}}, we analyze the performance of our network through gradient based analysis, investigate its generalization to different levels of noise, and compare its performance to different combinations of the segmenting and denoising tasks; 
    In \textbf{Section~\ref{sec:summary}} we summarize our conclusions and outline directions for future work.
    
    % \begin{itemize}
    %     \item 
    %     \item We design and train a convolutional neural network 
    %     \item We provide a meta analysis of the network's hyper-parameters and show that optimal performance is achieved when certain pre- and post-processing steps are applied to the network's output and data. Specifically, spatially smoothing out the target label maps improves the performance stability.
    %     \item 
    %     \item We : segmenting the noisy data directly; denoising and then segmenting; denoising and segmenting simultaneously, and show the superiority of segmenting directly.
    % \end{itemize}

    %  discusses the design of the training and validation data sets using TEM data of nanoparticles. Section 3 describes the network architectures considered. Section 4 describes the simulation results and their analysis, as well as results on real data. Section 5 describes the conclusions and suggestions for future work.
    \section{Related work} 
%CNNs for extracting information from image data have been extensively deployed in natural images for different tasks. Object detection based systems based on neural networks use either semi or fully convolutional architectures, to annotate object of interest in an image, providing a label and bounding box; In semantic segmentation, the network provides a map that assigns a different label to every pixel, either background or one of possible object classes, implicitly providing segmentation and classification.

\label{sec:related_work}

\subsubsection*{Deep Learning in electron microscopy}
    
%    CNNs have been successfully applied to denoising TEM images using a simulation-based dataset in \cite{Mohan2020Robust}, where it was shown that a large field of view is necessary to achieve state-of-the-art performance on TEM data. Deep CNNs have been applied to other image processing tasks in TEM beyond denoising \cite{muto2020application,spurgeon2021towards,lin2021temimagenet}. For example,~\cite{suveer2019super} proposes a CNN-based method for TEM image super-resolution, where CNNs are trained on pairs of low- and high-resolution images acquired experimentally.
% Refs.~\cite{khadangi2020stellar, roels2019cost} apply CNNs to perform semantic segmentation of real EM images in biological domain training on hand-labeled data. See Ref.~\cite{ede2020deep} for a comprehensive review.

Recently, deep learning has been applied to denoise TEM images using supervised learning based on simulated data~\cite{mohan2022deep,vincent2021developing}, unsupervised learning~\cite{udvd}, and semisupervised learning~\cite{mohan2021adaptive} (see \cite{mohan2025learn} for an overview tutorial). Beyond denoising, deep neural networks have also been employed for a range of other image processing tasks in TEM \cite{muto2020application, spurgeon2021towards, lin2021temimagenet}. For instance, \cite{suveer2019super} introduced an  approach for super-resolution of TEM images, where a neural network is trained on experimentally acquired pairs of low- and high-resolution images. In the biological domain, deep learning has been applied to perform semantic segmentation of real electron microscopy images using hand-labeled training data \cite{khadangi2020stellar, roels2019cost}. For a comprehensive overview of deep learning applications in electron microscopy see \cite{ede2020deep}.

    \subsubsection*{Depth estimation}
    %CNNs have been used extensively for information extraction in natural images \cite{lecun2015deep}. By far, the application that propagated furthest than any other is images classification, where a single label is to be extracted\cite{rawat2017deep}. Following applications in natural images, CNN based image classification has become standard in medical and scientific images as well. 
%  Information extraction is of great importance in other forms of image data as well. However, there are qualitative and quantitative differences when considering feature extractions in natural and scientific images. First, in natural images the number of features in an image ranges from 1 in image classification to a few dozens in most applications involving object detection and semantic segmentation. Scientific images, especially in microscopy can have many-fold more objects, making commonly used architectures computationally expensive both in runtime and memory requirements. Another challenge is the presence of noise in raw scientific images, necessitating a particular attention to the role of the noise in the feature extraction process, preventing the use of feature extraction algorithms that manipulate the input pixel value, such as cropping and resizing. 
    Depth estimation in natural images aims to recover 3D information from their 2D projections. Early approaches estimated 3D structure from multiview images by minimizing energy-based cost functions \cite{newcombe2011dtam} or by leveraging probabilistic priors \cite{pizzoli2014remode}. More recently, learning-based methods have achieved state-of-the-art performance in this domain. For example, DeMoN \cite{ummenhofer2017demon} trains a neural network to estimate both motion and depth from image pairs, while DeepMVS \cite{huang2018deepmvs} uses multiview images to predict pixel-wise depth for a reference frame. Estimating depth from a single image remains a more difficult task, due to the inherent ambiguity in recovering continuous depth values. To address this, \cite{cao2017estimating} reformulated the problem as semantic segmentation by discretizing depth into bins and training a residual CNN to predict the discretized map. Recent developments in autonomous driving have also driven interest in depth estimation as a critical component of object detection systems \cite{park2018high}. 
     \subsubsection*{Depth estimation in TEM}
     Electron tomography consists of the acquisition of multiple TEM images from different perspectives by rotating the object, with a subsequent processing to reconstruct the whole volume. 
     %Advances in the acquisition and reconstruction algorithms allow us to recover 3D structural information with atomic resolution. However, high-resolution scanning transmission electron microscopy (STEM) and TEM coupled with image simulations make possible the determination of a close 3D atomic arrangement by comparing experimental and simulated intensities. 
     Alternatively, scanning transmission electron microscopy (STEM) and TEM images can be used to estimate 3D atomic arrangements from a single image by comparing experimental and simulated intensities. In the case of high-angle STEM modes, the incoherent nature of the image gives rise to an intensity that is proportional to the thickness and the squared atomic number, allowing more straightforward inference of depth information~\cite{xu2015three,padgett2018connecting,zhou2019observing}. 
     %Alternatively, STEM or TEM images can be used to estimate 3D atomic arrangements by comparing experimental and simulated intensities. In the particular case of high angle annular dark-field STEM imaging, the incoherent nature of the image gives rise to an intensity which is proportional to the thickness, allowing a more straightforward inference of 3D information from a single 2D projection \cite{li2008three}. 
     However, all of these techniques require the acquisition of images with a high signal-to-noise ratio (SNR) to ensure accurate estimation. In general, SNR is inversely proportional to time resolution, which means that current approaches often cannot retrieve dynamic 3D information from data acquired at high resolution.

More recently, neural networks have been employed to estimate sample thickness in high-resolution TEM data.
\textcolor{black}{In \cite{madsen2018deep}, a segmentation network was developed to determine the number of atoms per column using simulated TEM images of metal nanoparticles, showing strong performance under high electron-dose conditions. However, the model was not validated on experimental data. In \cite{ragone2020atomic}, a regression-based network trained on simulated images was tested on real HR-TEM data of metal nanoparticles, with accuracy found to depend on the electron dose. In both studies, network performance was limited under low signal-to-noise ratio (SNR) conditions.} Our application to CeO2 nanoparticles presents additional challenges with respect to these works. The chosen crystal orientation reveals two distinct types of atomic columns: one composed of oxygen anions and the other of cerium cations. In phase contrast TEM images, these two columns have very different thickness-intensity dependencies and their occupancies  may be different due to crystal thickness effects or local variations in the oxygen vacancy concentration \cite{crozier2008situ,wang2009structural,wang2008measuring}.

     \subsubsection*{Semantic segmentation} Our approach reformulates depth estimation as semantic segmentation. This is a fundamental problem in computer vision, where the goal is to assign labels to pixels in an image indicating what entity they correspond to \cite{garcia2017review}. In natural images, these entities are typically objects, each occupying many pixels. Semantic segmentation networks function similarly to image classifiers but operate at the pixel level, identifying regions of interest and assigning labels based on spatial and contextual features within the image domain. Recent advances, such as transformer-based detection models \cite{carion2020end}, incorporate broader contextual information to improve performance. However, the most salient features in segmentation tasks often remain dominated by localized visual cues \cite{vinogradova2020towards}. 
\subsubsection*{Segmentation in the presence of noise} Semantic segmentation in the presence of noise is a critical challenge in both medical and scientific imaging applications \cite{asgari2021deep}. In \cite{buchholz2020denoiseg}, it was proposed that jointly training a network to perform both denoising and segmentation can improve results. However, we did not observe this in our setting; joint denoising and segmentation did not improve performance. 

\section{Methodology}
\label{sec:methodology}
\subsection{The SegDepth Framework}
\label{sec:framework}
%Ideally we would like to estimate the three-dimensional structures of nanoparticles. Nanoparticles are ordered in preferential directions, so when observing the particle from a zone axis (high-symmetry orientation), all atoms are ordered in straight columns, with an atomic column depth associated with each column. We can thus transform the problem to an atomic depth estimation.

We propose an approach to infer the depth of atomic columns of nanoparticles ordered in preferential directions from TEM images acquired from a zone axis (high-symmetry orientation). To estimate the depth, we cast this task as a semantic segmentation problem, where the labels encode the number of atoms in the atomic column corresponding to each pixel (or indicate that the pixel belongs to the background). This segmentation task is addressed by training a convolutional neural network to estimate the labels. 

% The proposed network is as follows: 
% Given two dimensional TEM image, $\mathbf{x}$   $$ \mathbf{x}\in\mathbb{R}^{M_1\times M_2},$$  the network outputs  a depth mask, $f$, 
% $$f_\theta(\mathbf x)\in\{C+1\}^{\alpha M_1\times \alpha M_2},\hspace{1em} \alpha<1$$ such that where every pixel takes an integer value in $0,1,\dots C$, corresponding to the depth of the atomic column the pixel is part of (0 is reserved for the background). We propose a simulation-based training set of images and depth masks to train the network.

% State-of-the-art techniques for information extraction in natural image and video data rely on images that are annotated for object labels and/or their locations. The main challenge is determining persistent features that can be used to segment different regions in the image and their values. This methodological approach is impractical when considering electron microscopy data, which may be acquired at high temporal resolution. At these scales, the ground truth cannot be inferred. Furthermore, a high temporal resolution implies a lower signal-to-noise ratio, making the information extraction all the more challenging. Thus, the most viable option is to design a simulation-based training data set, for which the full information of the underlying structure is known. 

As depicted in Figure~\ref{fig:scheme} the SegDepth neural network is trained using simulated data. %Similar to \cite{mohan2020deep}, we use a simulated based training methodology. 
Using prescribed three dimensional atomic structures, we generate synthetic TEM images and corresponding segmentation maps, encoding the column depth at every pixel. 
%$$\mathbf{d}\in \{C+1\}^{M_1\times M_2},$$ which are target semantic segmentation labels, used during training and validation.
In order to mimic shot noise present in real data, the TEM images are corrupted using Poisson noise. Once trained on the synthetic data, the network can be applied to estimate depth in real TEM data. %training, the TEM images are corrupted by simulated Poisson noise and then fed into a fully convolutional neural network, whose output is then matched to the target mask. In validation, the performance of the network is evaluated against simulated images that were not used in training (see Figure~\ref{fig:scheme} for an overview of the methodology).

%There are several crucial design aspects to the construction of the segmentation network, which include:
%Design of the simulation-based training dataset, including: Generating simulated TEM images; Depth maps for each nanoparticle structure; Training CNNs for segmentation on noisy samples of TEM image as input and depth maps as output.

%In the following subsections we explain each step in detail.

\subsection{Data simulation}
\label{sec:sim_dataset}

\subsubsection{Nanoparticle atomic models}
{\textcolor{black} {The simulated data are based on atomic models of nanoparticles that resemble the nanoparticles in the real data of interest. In our case, we considered a cerium oxide CeO2 (110) surface viewed along the [110] crystallographic zone axis orientation, obtained using the freely available Rhodius software \cite{rhodius2025,rhodius2}. The surface of the atomic models have different geometric features, i.e., flat, sawtooth, or stepped shapes,  previously observed in the experimental data. The average sample thickness ranged from 3 to 10 atoms. In addition, a gradual decrease in the depth of the atomic column near the surface was included, simulating the wedge structures of real nanoparticles close to their surface.}}

\subsubsection{TEM images}
{\textcolor{black} {TEM images were generated from the nanoparticle atomic models via the multi-slice approach, implemented using the Dr. Probe software package~\cite{barthel2018dr}. 
%Various hyperparameters that control the morphology of the sample were modified throughout the datasets in order to train the network, and described in the following paragraphs. These variations generate significant modulations in the intensity of the TEM image, which are likely to vary during a regular TEM experiment.
}}
{\textcolor{black} {
%In addition to variations in the atomic structure of the sample, different TEM 
Variations in electro-optical parameters during typical TEM experiments give rise to significant modulations in the intensity of the resulting TEM images. In order to account for this, images were generated using different parameters of the electron microscope. For example, the range of the defocus value (which is critical, as it is likely to change during a real experiment) was set between 1 and 9 nm.}} %We have generated in total 864 image simulations from 96 CeO2 (110) surface wedge models by systematically changing imaging conditions and structural configuration, for example, defocus, thickness, etc. 
To match the experimental data, the third order spherical aberration coefficient (C3) was set to -9 $\mu$m resulting in white oxygen column contrast in the simulation \cite{urban2009negative}, and the fifth order spherical aberration coefficient (C5) was set to be 5 mm. Lower-order aberrations were set to be 0 as those parameters are constantly tuned to be near negligible value during experiments. 

% {\textcolor{black} {The variations in structural arrangements (surface geometry type and gradient toward the surface), atomic depth, and defocus value result in a dataset of CeO2 (110) surface TEM simulated images, used for training and testing CNN.}}

We note that there are some other parameters that are less likely to be modified during a particular TEM experiment, which have been kept unchanged. For instance, images were simulated with 2048 x 2048 pixels and then down-sampled by a factor of 2 to match the approximate pixel size of the experimentally acquired image series, that is, 5.3 pm/pixel. Other aberrations, such as C3 and C5, were set to -0.009 mm and 5 mm, respectively, similar to experimental values.

\subsubsection{Noise} 
The synthetic TEM images are corrupted by Poisson noise, simulating the noise produced by the behavior of low-dose electron beam. At each pixel, the value of the noisy TEM image $x$ is generated by sampling independently from a Poisson distribution with parameter $\lambda c$ and scaling the result:
\begin{align}
x \sim\frac{1}{\lambda}\text{Poisson}(\lambda c),
\label{eq:noise_simulation}
\end{align}
where $c$ is the value of the clean TEM image at that pixel and $\lambda$ is a parameter that determines the noise level. Since the mean and variance of a Poisson random variables are both equal to the parameter, by linearity of expectation the ratio between the mean and standard deviation of the signal at each pixel is proportional to $\sqrt{\lambda}$:
 \begin{align}
 \frac{\mathbb{E}(x)}{\text{std}(x)} = \frac{\lambda c / \lambda}{\sqrt{\lambda c / \lambda^2}} = \sqrt{\lambda c}. \label{eq:lambda}
 \end{align}
 Consequently, larger values of $\lambda$ correspond to higher signal-to-noise ratios.
% signal-to-noise ratio (SNR) is equal to the ratio between the mean and the standard deviation of the noise. Since the mean and variance of a Poisson random variables are both equal to the parameter, by linearity of expectation the SNR is 
% \begin{align}
% \text{SNR} = \frac{\lambda c / \lambda}{\sqrt{\lambda c / \lambda^2}} = 
% \end{align}
%At each pixel, the value of the noisy TEM image is generated by sampling independently from a Poisson distribution with parameter $\lambda x$, where $x$ is the value of the clean TEM image at a pixel of the TEM image noisy image random variable $Y$ is generated from the clean image $\mathbf x$ by $Y(\mathbf x)\sim\frac{1}{\lambda}\text{Pois}(\lambda \mathbf x)$. Independent samples are then drawn from the distribution. 
%The parameter $\lambda$ affects the variance
% \begin{equation}
%     \mathbb{E}[Y(\mathbf x)]=\mathbf x,\hspace{1em}\text{Var}(Y(\mathbf x))=\frac{\mathbf x}{\lambda},
%     \label{eq:pois_scale}
% \end{equation}
% so we can control the noise level: ratio of mean and standard deviation of a Poisson random variable is 1. The weighting by $\lambda$ in the matter described modifies it to $\sqrt{\lambda})$.
\subsubsection{Depth segmentation maps}
\label{sec:depth_segmentation_maps}
\begin{figure}[t]
        \centering
        \begin{tabular}{c c c }
        Depth mask& TEM simulation & Noisy input\\
            \includegraphics[width=0.3\linewidth]{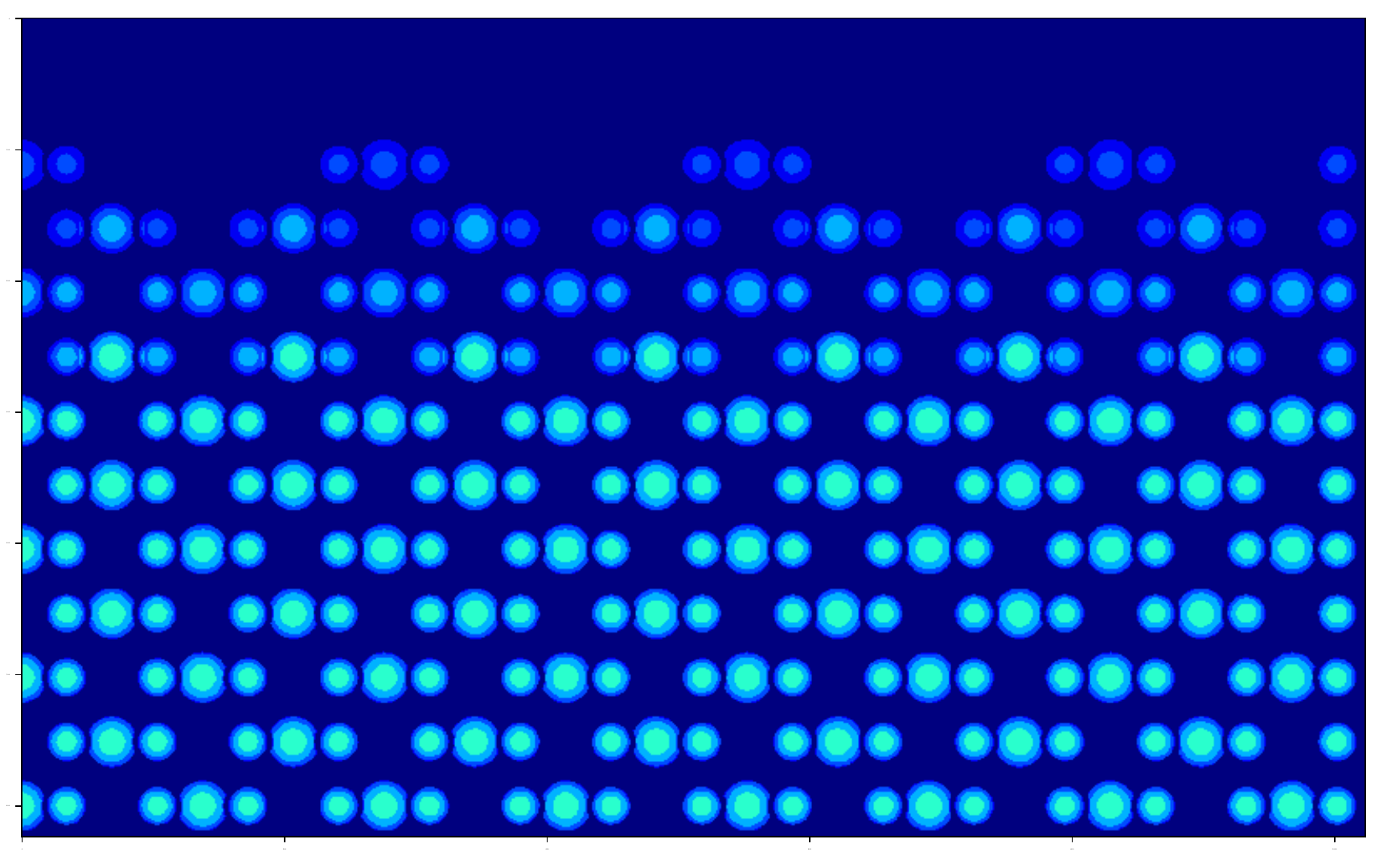} & \includegraphics[width=0.3\linewidth]{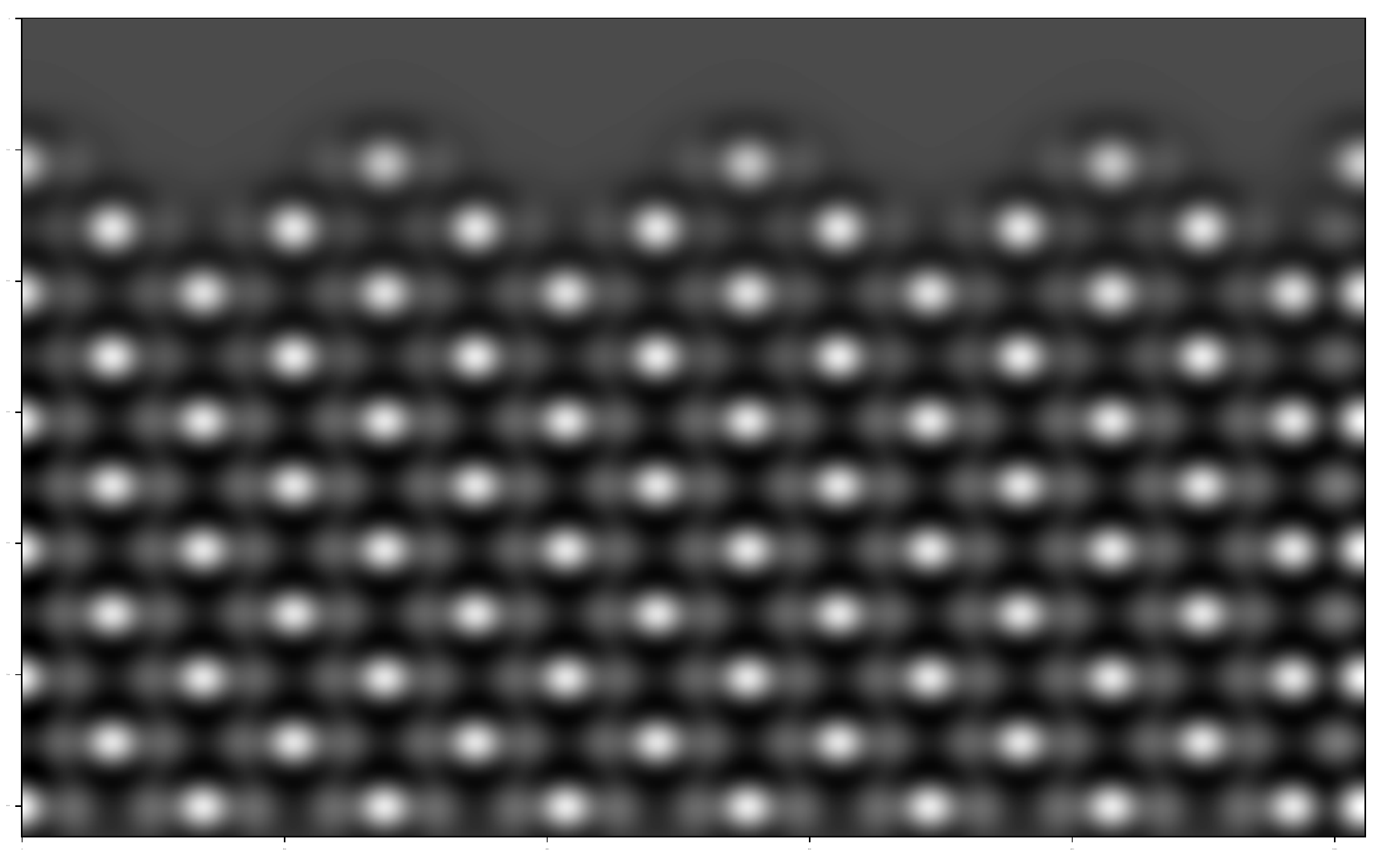} & \includegraphics[width=0.3\linewidth]{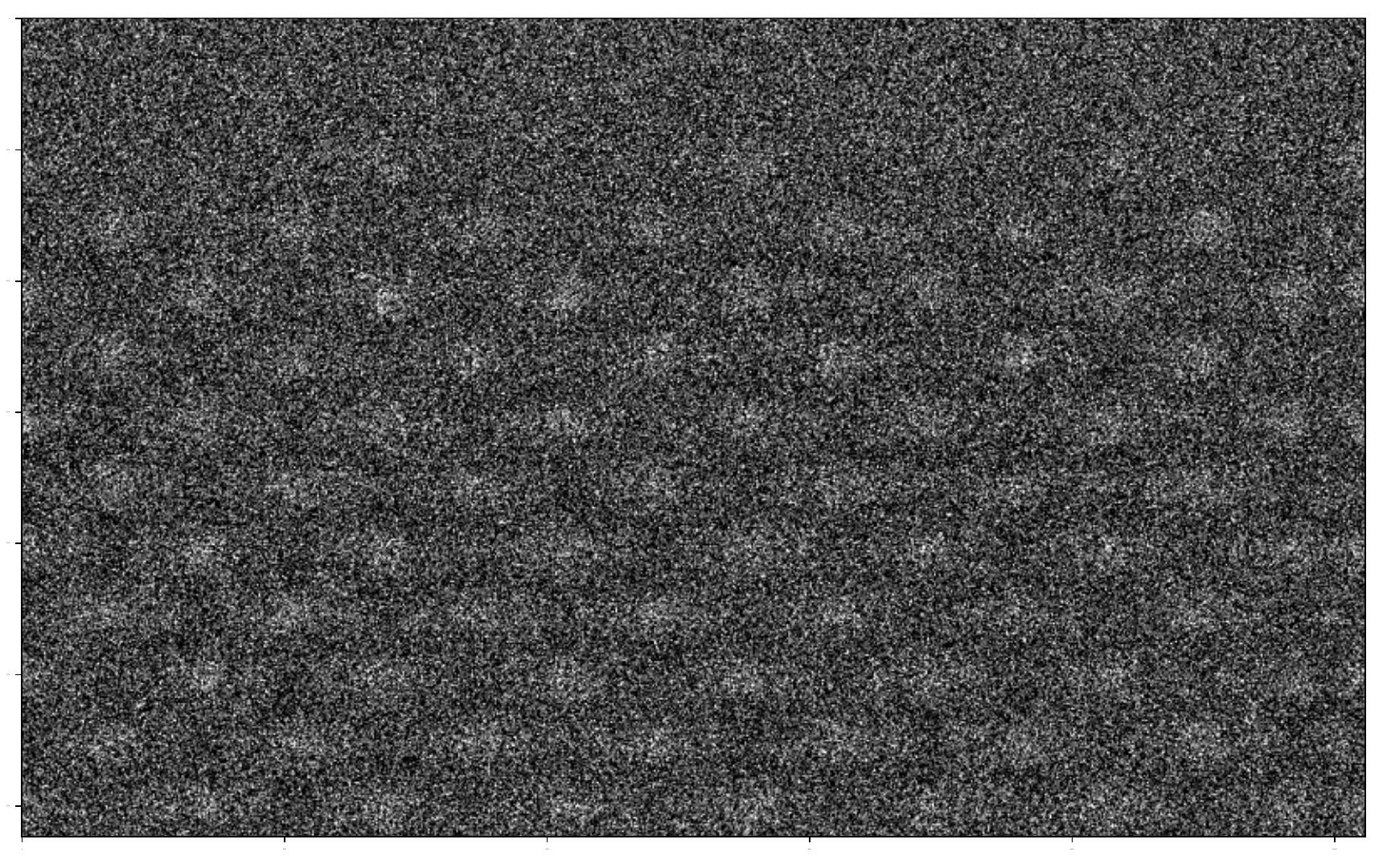}\\
            \includegraphics[width=0.3\linewidth]{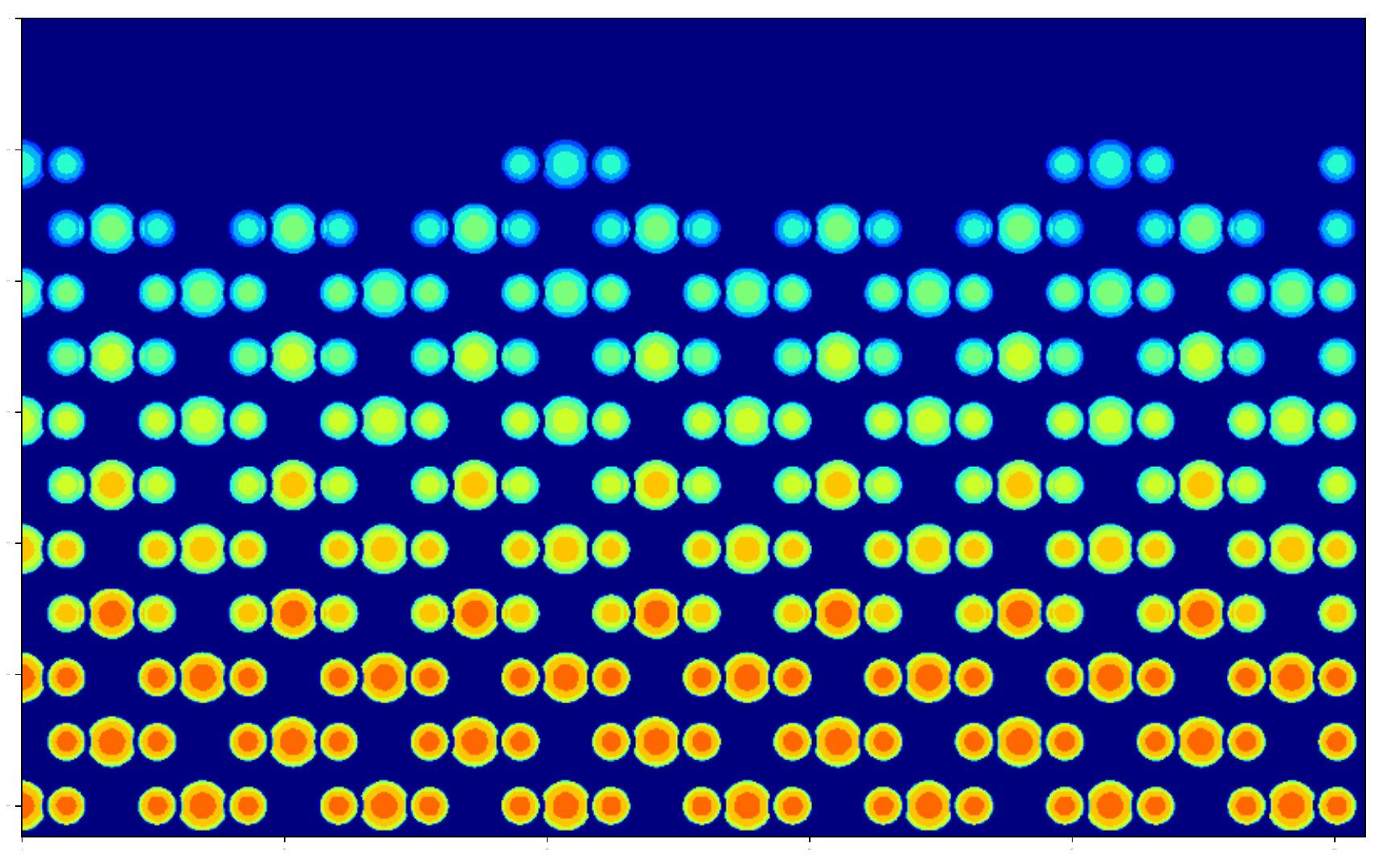} & \includegraphics[width=0.3\linewidth]{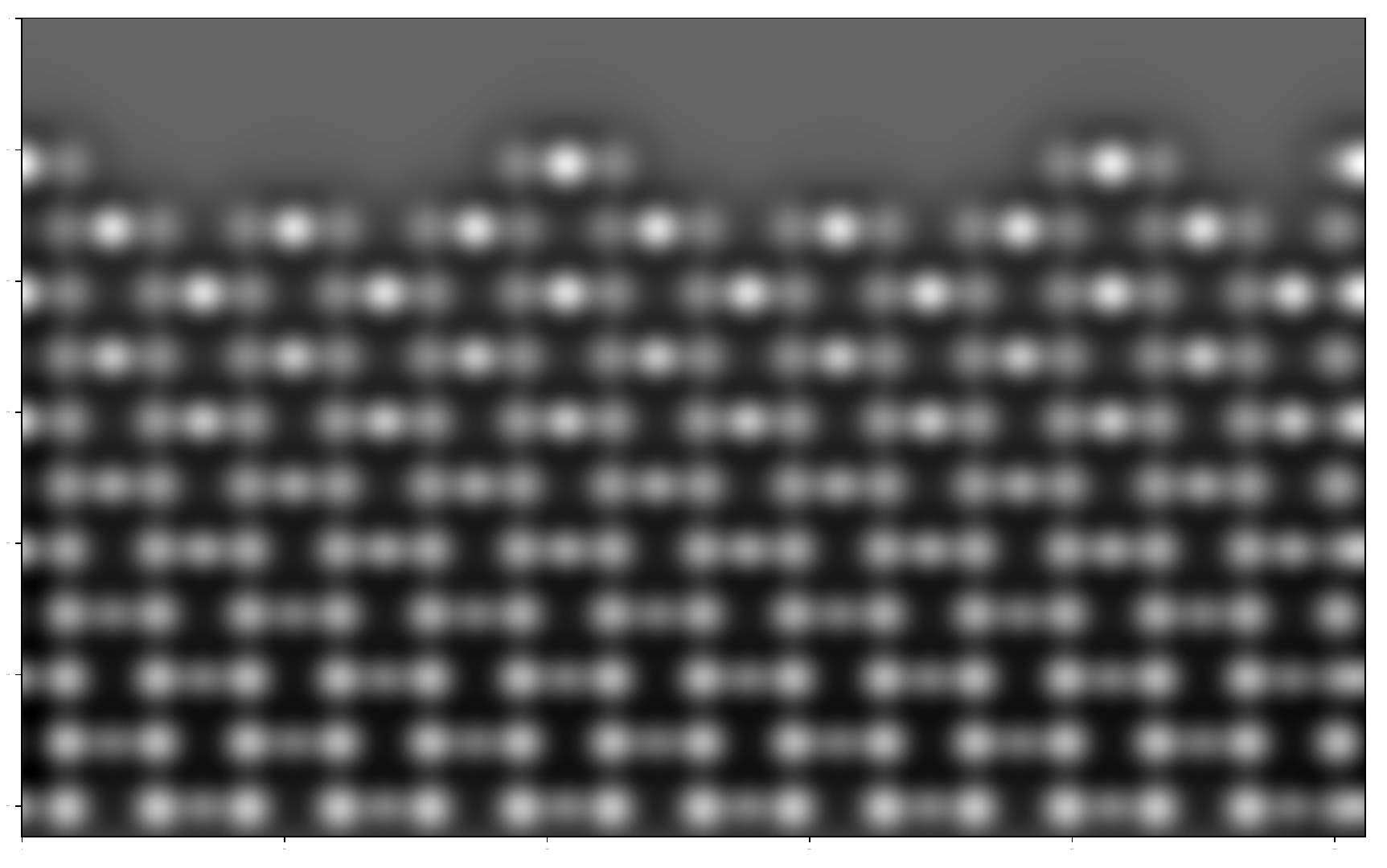} &
            \includegraphics[width=0.3\linewidth]{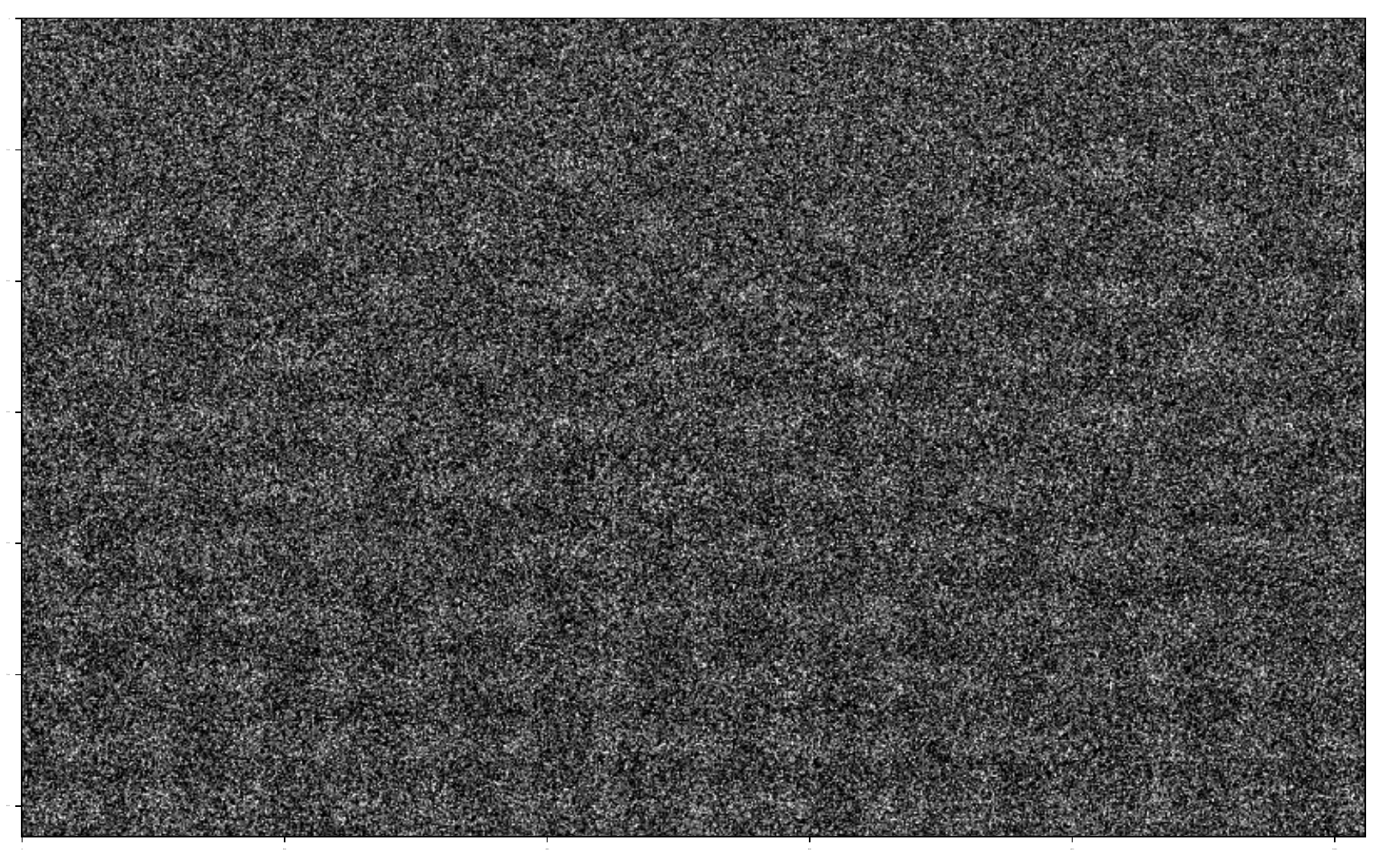}\\
            \includegraphics[width=0.3\linewidth]{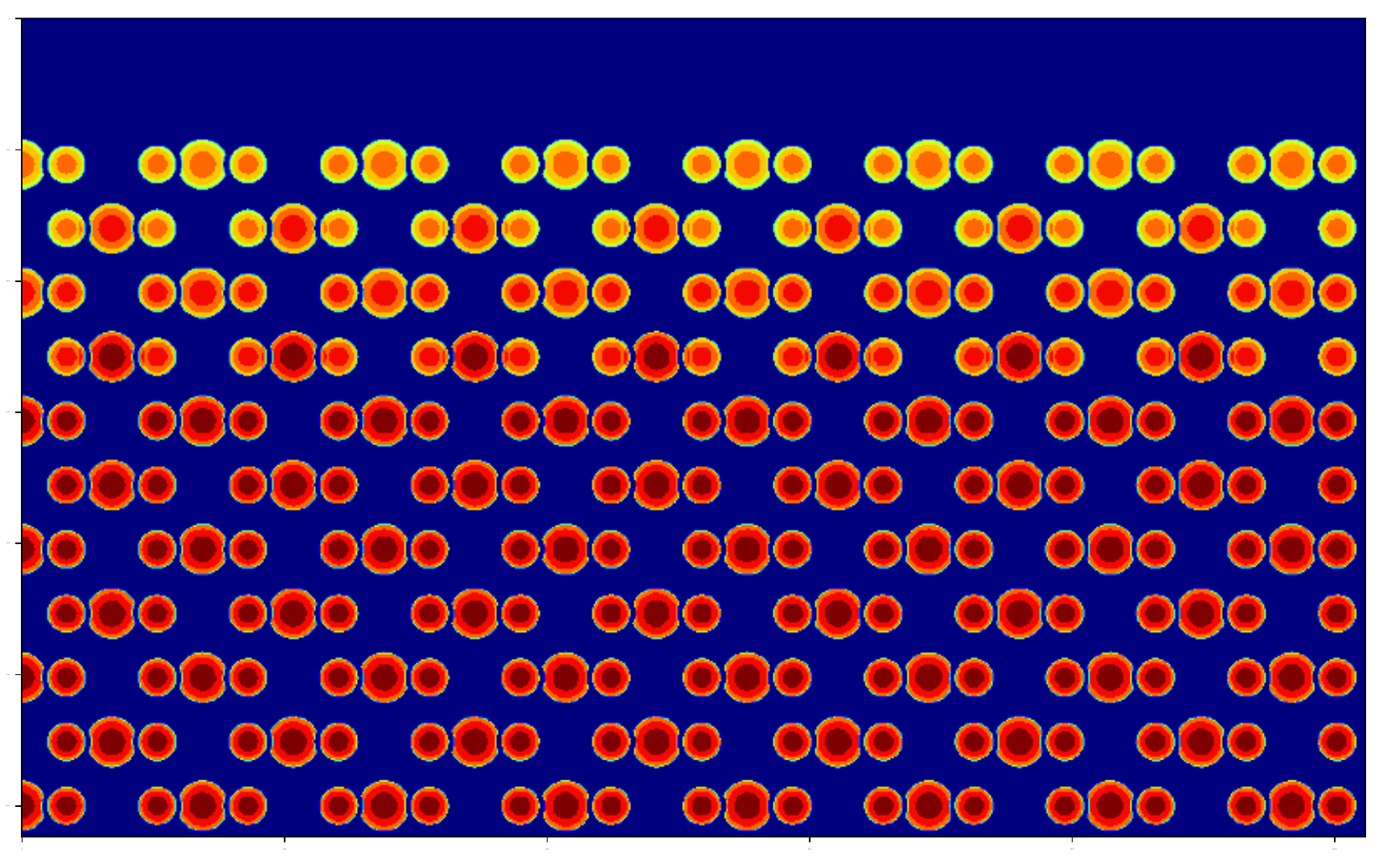} & \includegraphics[width=0.3\linewidth]{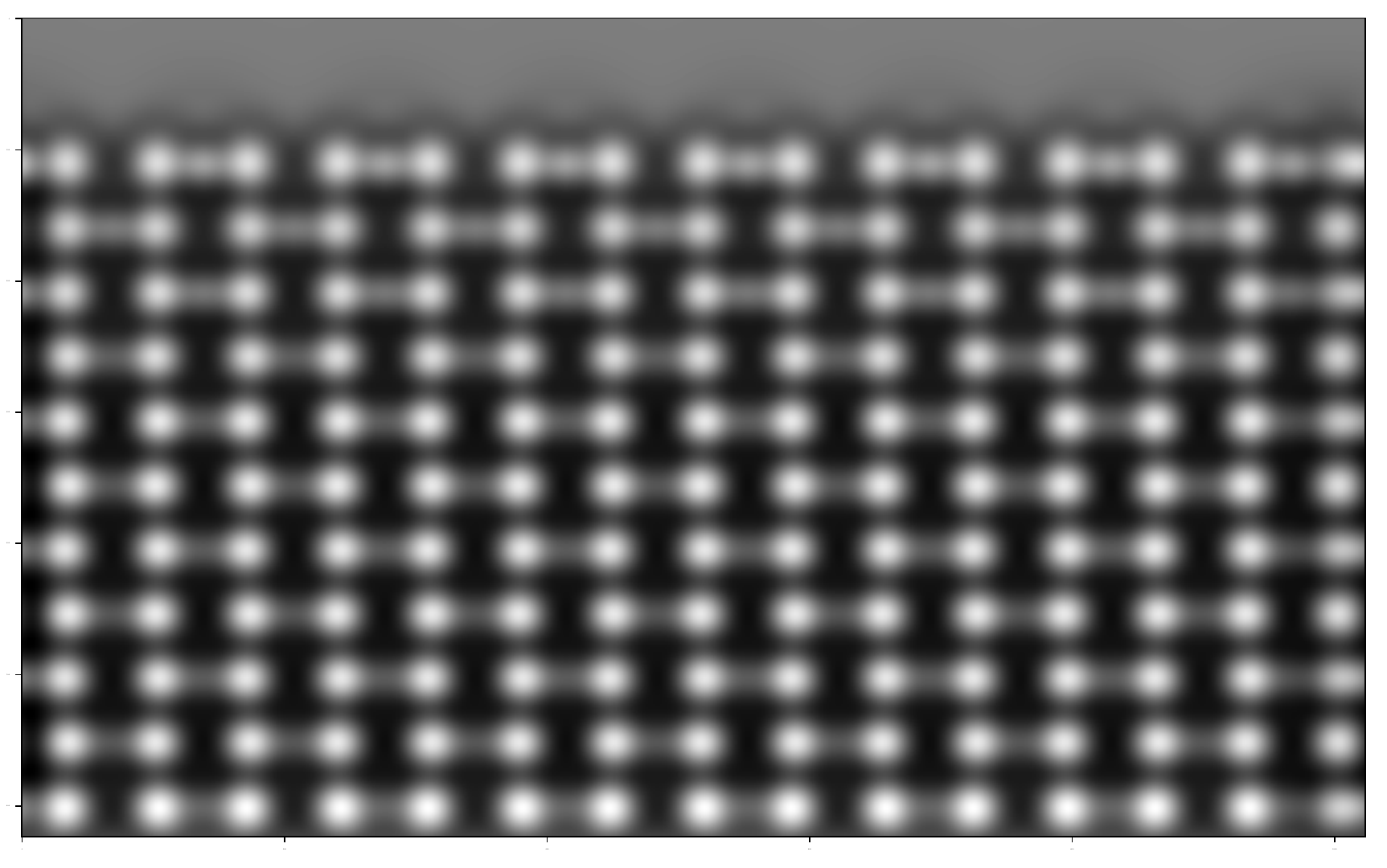} & \includegraphics[width=0.3\linewidth]{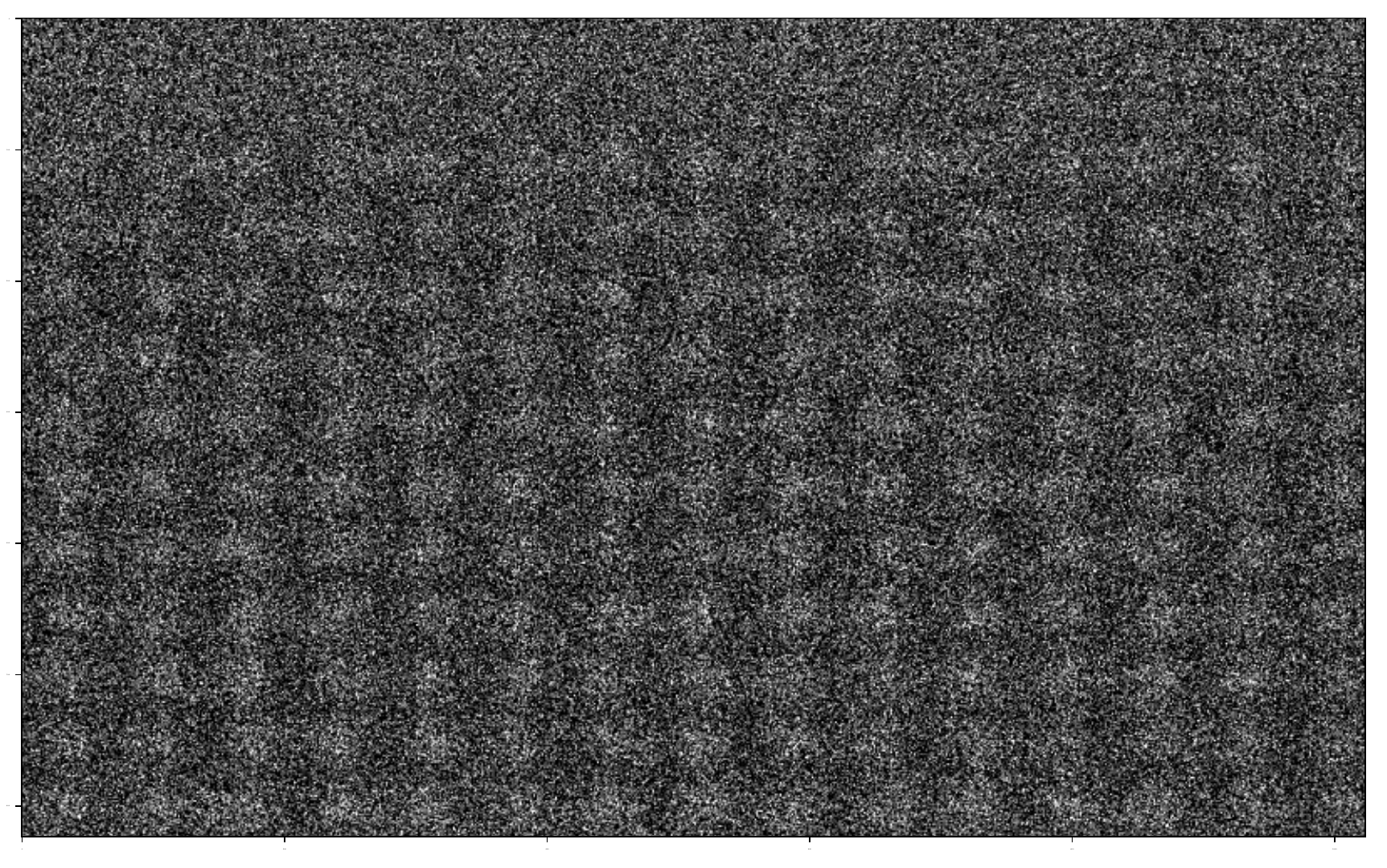}\\
            \multicolumn{3}{c}{\includegraphics[width=0.48\linewidth]{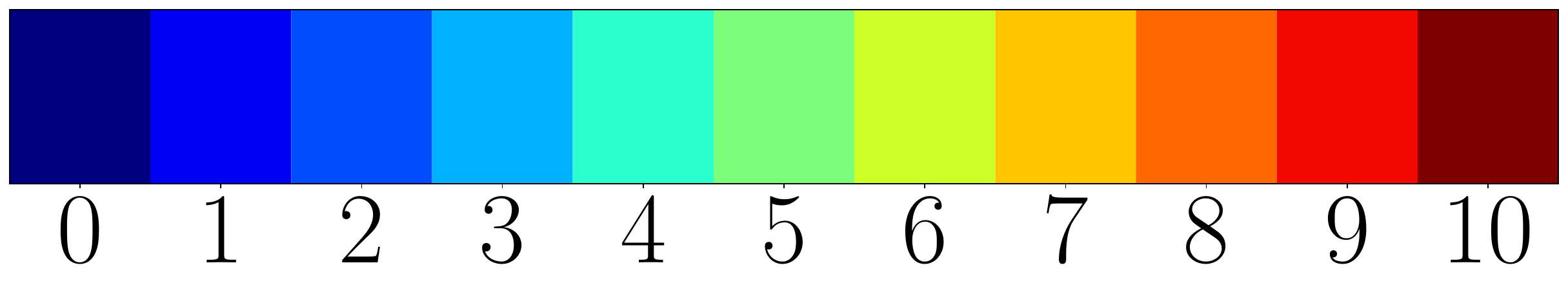}}
        \end{tabular}
        \caption{\textbf{Data simulation.} Examples of simulated depth masks (left) and their associated TEM images before (center) and after (right) corrupting them with Poisson noise. 
        %Structures are represented by two-dimensional projections of the atomic column depth. 
        %Structures can vary in both depth and the geometry of the surface. 
        % The depth of the atomic column greatly affects the contrast of different elements in the nanoparticle. Actual intensity measurements follow Poisson statistics, limiting the data available for inference.  [need labels on figures (a,b, c ) an d]
        }
        \label{fig:structures}
    \end{figure}

In order to generate the depth segmentation map associated with each TEM image, we compute the depth of the atomic column corresponding to each pixel from the three-dimensional  structure of the nanoparticle atomic model used to simulate the TEM image. This is achieved by projecting each atom onto the imaging plane, and counting how many are mapped to each pixel. Examples of segmentation maps corresponding to different TEM images are shown in Figure~\ref{fig:structures}. If no atoms are projected on a particular pixel, it is considered part of the background and assigned a depth of 0.

   \addtolength{\tabcolsep}{-4pt}  
    \addtolength{\tabcolsep}{4pt} 

%leverage three-dimensional information, encoded in the crystallography structure file used to generate the TEM simulation. The file contains information about the three-dimensional positioning of atoms in the particle, represented as 3D coordinates in a normalized supercell, with the same axis orientation as the TEM image. We assign each atom a radius $r_k$, determined by the element and imaging setup, and then project the atoms onto the image plane. We denote the projected center coordinates $\mathbf e_k$. The number of atoms overlapping at a particular position determines the depth, $d$,  of the atomic column in that pixel, i.e.,
% \begin{equation}
%     \mathbf{d}=d(\mathbf x )=\sum\limits_k \mathbf{1}_{\{\| \mathbf x- \mathbf e_k\|\le r_k\}}
% \end{equation}
%The background is assigned a 0 value, and atomic column values go up to 10 for the simulated data. The coordinates $\mathbf{x}$ are chosen to correspond to the original dimensions of the image.

\subsection{Deep learning}
\label{sec:deep_learning}
In the proposed SegDepth framework, a deep neural network is trained to estimate the segmentation maps encoding atomic-column depth from noisy TEM images. We utilize a popular architecture for image segmentation, the UNet~\cite{unet}, which consists of convolutional layers, as well as downsampling and upsampling layers that enable the network to capture multiscale structure. UNets have been successfully applied to denoise TEM data~ \cite{mohan2025learn}. We follow a similar architecture to \cite{mohan2022deep}, which includes six downscaling layers. The architeture is described in the appendix.

The output of the UNet is fed into two downsampling layers, which produce an output with the same dimensions as the segmentation map. Then median filtering of kernel size 4 is applied to enforce similar estimates at adjacent pixels (which are more likely to belong to the same atomic column). Finally, the resulting values at each pixel are fed into a softmax layer, which generates an estimate of the probability that each pixel belongs to an atomic column with 0 to 10 atoms. Given an input image $y$, we denote the final output by $F_{\theta}(y)$, where $\theta$ represents the network parameters. For $1 \leq i \leq N$, where $N$ is the total number of pixels in $y$, $F_{\theta}(y)_i$ is a vector of dimension 11, which approximates the probability that the atomic column associated with pixel $i$ contains each possible number of atoms (from 0 to 10):
\begin{align}
\mathbb{P} \left(d \text{ atoms in pixel } i\right) \approx F_{\theta}(y)_i[d], \quad 0 \leq d \leq 10. 
\end{align}
The estimated depth $\hat{d}_i$ at the pixel is set equal to the entry with the highest probability, 
\begin{align}
\hat{d}_i := \arg \max_{0 \leq d \leq 10} F_{\theta}(y)_i[d]. 
\end{align}
The corresponding probability can be interpreted as the \emph{confidence score} $c_i$ of the model at that pixel~\cite{hendrycks2016baseline}, and can be used for uncertainty quantification,
\begin{align}
c_i := F_{\theta}(y)_i[\hat{d}_i], 
\label{eq:confidence}
\end{align}
as illustrated in the rightmost image of Figure~\ref{fig:example}.

\begin{figure}[t]
   \begin{tabular}{c c}
       \includegraphics[width=0.42\linewidth]{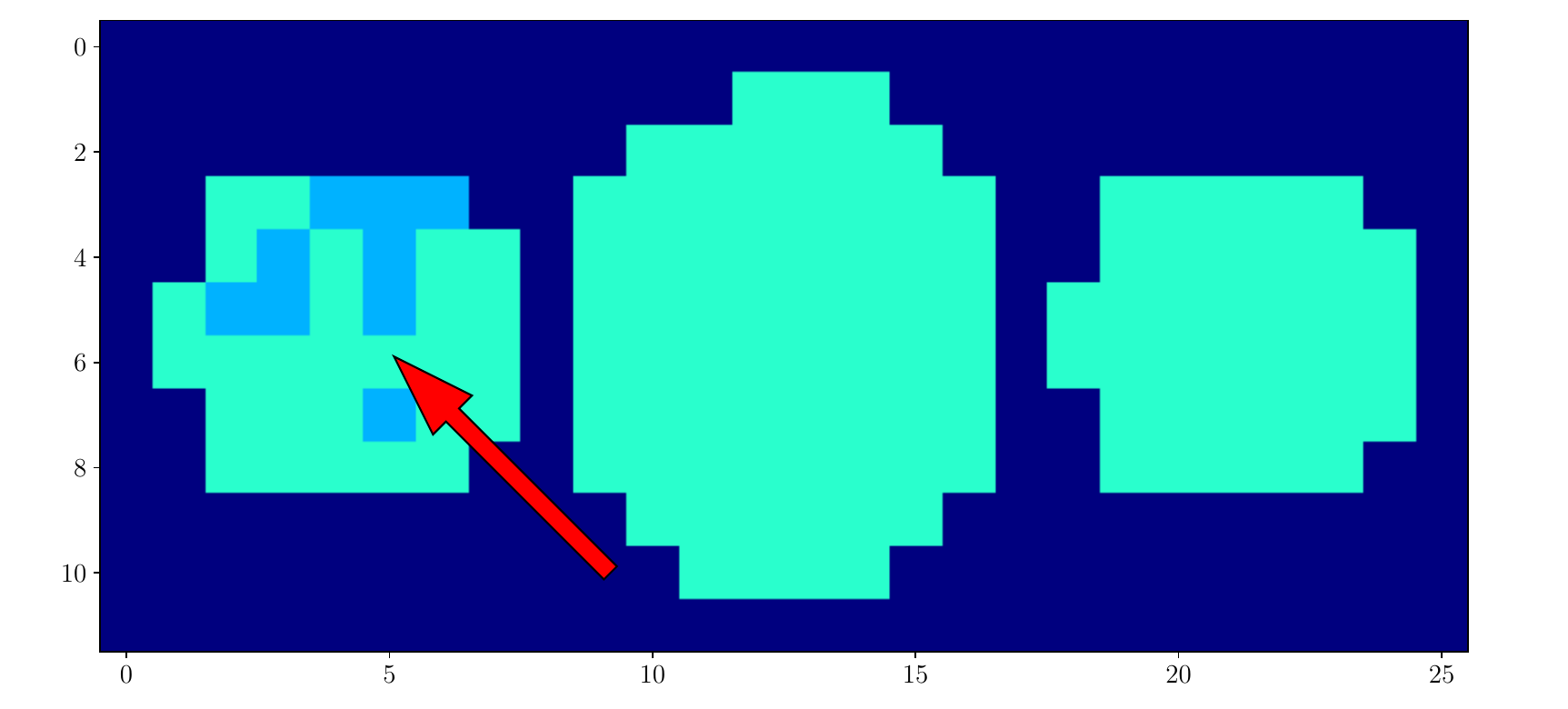}& \includegraphics[width=0.42\linewidth]{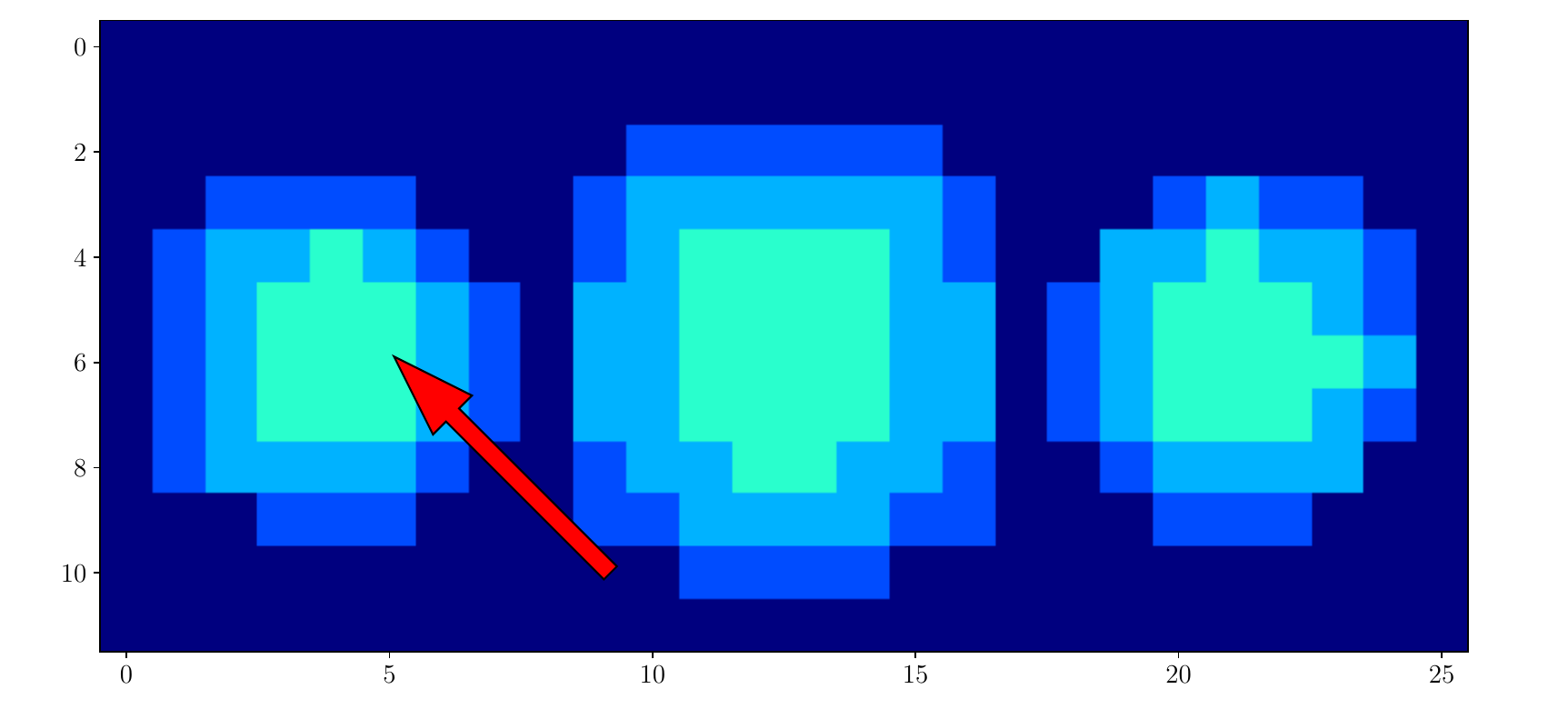}\\
       $(a)$ &$(b)$
   \end{tabular}
        \caption{\textbf{Spatial label smoothing.} (a) The pixel-wise depth segmentation labels obtained by projecting the nanoparticle atomic structure onto the imaging plane, as described in Section~\ref{sec:depth_segmentation_maps}, 
        contain irregular noisy patterns that
        change abruptly. (b) Spatial smoothing, so that the labels vary in concentric rings around each atomic column, produces more regular patterns that are easier to learn.} 
        \label{fig:smooth}
\end{figure}

In order to train the network, its output is compared to a \emph{smoothed} version of the segmentation map described in Section~\ref{sec:depth_segmentation_maps}. As illustrated in Figure~\ref{fig:smooth}, the pixel-wise depth labels, computed as described in Section~\ref{sec:depth_segmentation_maps}, contain irregular edges (e.g. changing abruptly from 0 in the background to the number of atoms in an adjacent atomic column), and can be noisy due to the projection onto the imaging plane. These patterns are difficult to learn and could result in overfitting. Consequently, we smooth the pixelwise labels, so that they decrease gradually from the atomic column to the background in concentric rings. This improves network performance and makes the interpretation of the estimated segmentation map less ambiguous. 

During training, we divide the training set of clean TEM images into batches. For each fixed batch of $B$ images $x^{[1]}$, $x^{[2]}$, ..., $x^{[B]}$, at each training epoch we corrupt the images with noise following \eqref{eq:noise_simulation}, to obtain the noisy TEM images $y^{[1]}$, $y^{[2]}$, ..., $y^{[B]}$. Generating fresh noisy images at each epoch, a technique known as continuous noise sampling~\cite{mohan2025learn}, prevents overfitting specific noise realizations. The network parameters are learned by minimizing a weighted cross-entropy cost function, equal to the weighted negative likelihood of the observed (smoothed) number of atoms at each pixel according to the deep neural network across the whole batch,
\begin{align}
\mathcal{L}(\theta):= - \sum_{j=1}^{B} \sum_{i=1}^{N} w_{i,j} \log F_{\theta}(y^{[j]})_i[s_i^{[j]}]
\label{eq:weighted_ce}
\end{align}
where $s_i^{[j]}$ is the number of atoms in the $i$th pixel of the smoothed segmentation map corresponding to the $j$th TEM image in the batch. 

\begin{figure}
    \centering
\includegraphics[width=0.7\columnwidth]{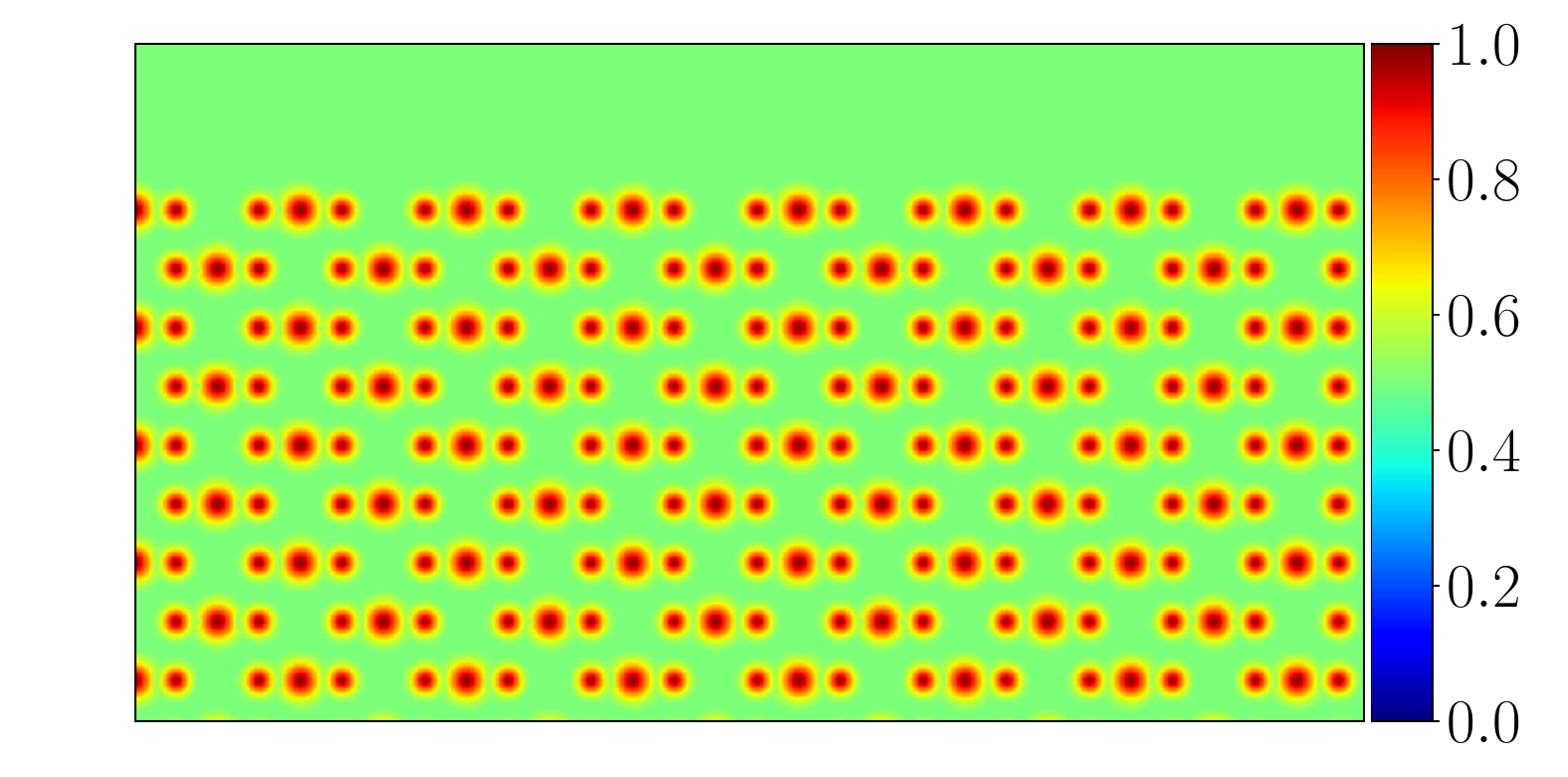}
    \caption{\textbf{Spatial weighting to address class imbalance.} We utilize a spatial weight within the training loss in \eqref{eq:weighted_ce} to penalize misclassification in pixels that are closer to the atomic column centers. This mitigates class imbalance due to the abundance of background pixels that do not contain atomic columns.%Since the dominating class is the background, unweighted loss functions lead to well segmented images, separating background and particle, but under perform in actual classification.
    }
    \label{fig:spatial_weights}
\end{figure}

Weighting is applied to the cross-entropy loss in \eqref{eq:weighted_ce} to  address class imbalance~\cite{buda2018systematic}, due to the fact that most of the pixels in the TEM images are background.
The weight $w_{i,j}$ assigned to pixel $i$ in image $j$ is equal to one at the center of each atomic column and decreases gradually away from it, as illustrated in Figure~\ref{fig:spatial_weights}. It  penalizes misclassification of pixels belonging to atomic columns more strongly than those in the background. 

In order to train the deep neural network, the loss~\eqref{eq:weighted_ce} is minimized with respect to the network parameters $\theta$ via stochastic gradient descent using the Adam optimizer with learning rate scheduling~\cite{kingma2014adam}. The training set consists of $8.4\times10^3$ simulated TEM images and their corresponding smoothed segmentation maps. Early stopping is performed by minimizing the loss on a held-out validation dataset with $3.6\times 10^3$ TEM images.

\begin{table*}[t]
    \caption{Performance of the proposed framework on the test set of simulated TEM data for different sizes of the underlying neural network model.}
\centering
\begin{tabular}{cccccc}
        \toprule
        Base& \# parameters &Pixelwise&Center&Real Atom &Hallucinated Atom\\
        Channels&$(\times 10^6)$&Accuracy&Accuracy&Detection Rate&Rate ($\times 10^{-2}$)\\
        \midrule
        16 & 11.75&0.931&0.812&0.955&7.25\\
        32&46.92&0.932&0.842&0.944&3.58\\
        64 &18.75&\textbf{0.939}&\textbf{0.847}&\textbf{0.956}&2.33\\
        96&42.17&0.936&0.83&0.943&3.62\\
        130&77.31&0.888&0.699&0.939&\textbf{1.75}\\\bottomrule
    \end{tabular}
    \label{tab:net_acc}
\end{table*}

\section{Experiments with Simulated Data}
 \label{sec:simulated}
In order to evaluate the performance of the proposed SegDepth framework, we apply it to a held out test data set of noisy TEM images paired with the corresponding ground-truth depth segmentation maps, generated as explained in Section \ref{sec:sim_dataset}. 

\subsection{Metrics}
We use the following four metrics to quantify to what extent the framework is able to recover the atomic columns and estimate their depth:
\begin{itemize}
    \item \textbf{Pixelwise accuracy:} Fraction of pixels where the estimated depth is correct.
    \item \textbf{Center accuracy:} Fraction of pixels, within a neighborhood of the center of an atomic column, with radius of 16 pixels for Ce atoms and 12 pixel for O atom, where the estimated depth is correct. This metric is motivated by the fact that most pixels belong to the background, which inflates the pixelwise accuracy, whereas we are mostly interested in the pixels corresponding to atomic columns.
    \item \textbf{Real atom detection rate:} Fraction of pixels belonging to atomic columns, which are detected (the estimated depth is greater than zero).
    \item \textbf{Hallucinated atom rate:} Fraction of pixels estimated to belong to atomic columns (the estimated depth is greater than zero), which correspond to the background.
\end{itemize} 
\subsection{Model size}
\begin{figure}[t]
    \centering
       \includegraphics[width=0.7
       \columnwidth]{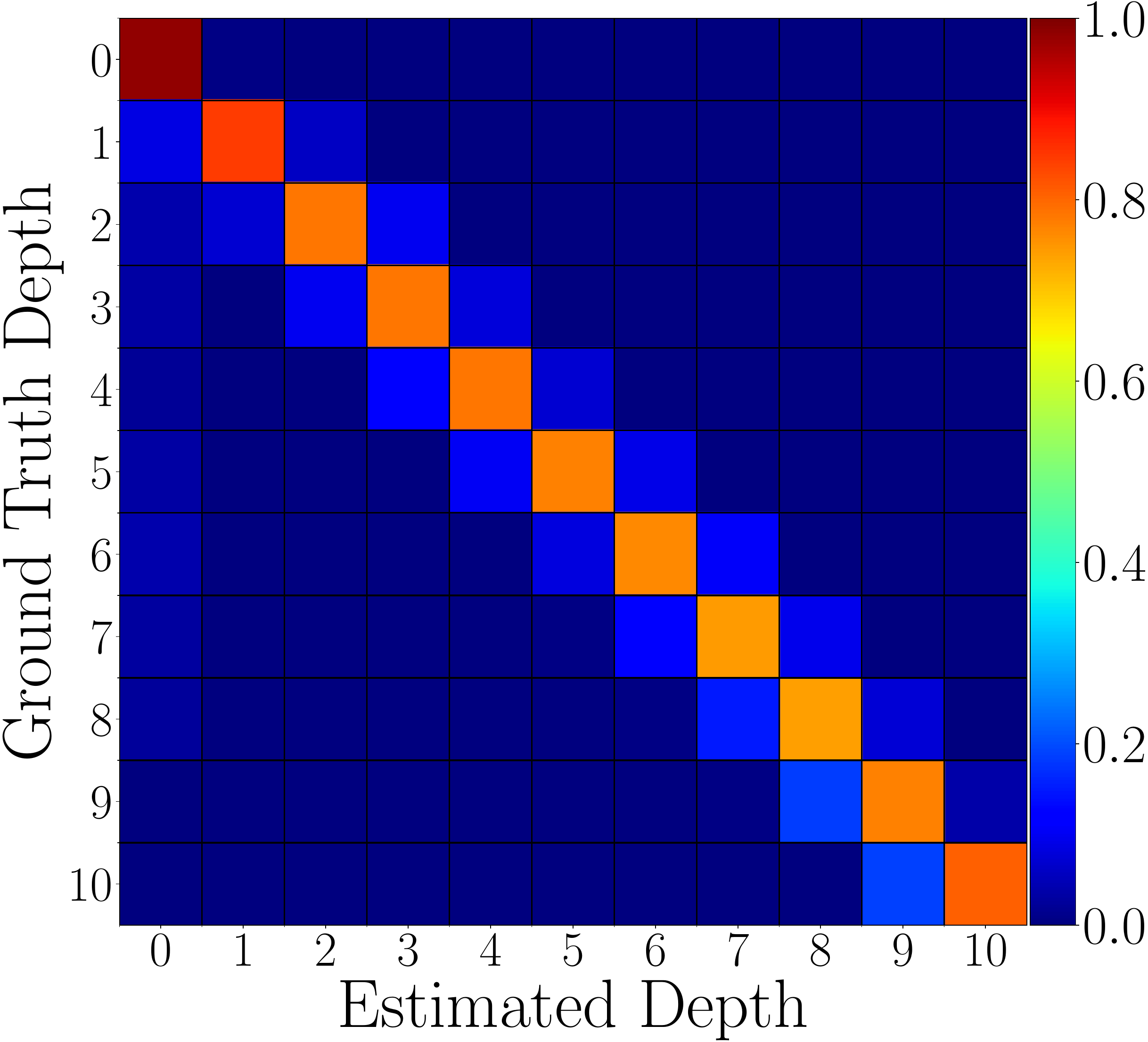}
\caption{\textbf{Confusion matrix.}: The confusion matrix compares the  ground-truth depth and the depth estimated by the proposed framework on the test dataset of simulated TEM data, when the noise level parameter is set to 1, and the number of baseline channels equals 64. The confusion matrix is approximately diagonal, showing that the estimated depth is mostly within $\pm 1$ atom of the ground truth depth.
%$CM_{\ell,\ell'}$ is the fraction of pixels with true label $\ell$ classified as $\ell'$. As the network converges, the confusion matrix trends toward a diagonal matrix. A pixel predicted atomic column rarely deviates by more than $\pm1$, giving an extended degree of confidence to the network's predictions. 
} 
        \label{fig:conf_mat}
\end{figure}

Table \ref{tab:net_acc} reports the evaluation metrics for UNet architectures with different numbers of base channels.\footnote{In a UNet architecture the base channels are the number of channels in the first convolutional layer of the encoder. Then the number of channels doubles with each downsampling layer.} The noise parameter $\lambda$, which determines the SNR as justified by \eqref{eq:lambda}, is set to 2.5. The largest architecture underperforms the rest, particularly in terms of center accuracy, suggesting that it may overfit the training data to some extent. The remaining architectures achieve high pixelwise accuracy, center accuracy and real atom detection rate, while having a small hallucinated atom rate. This establishes that the framework is able to recover the underlying atomic structure from the TEM images and accurately estimate the atomic-column depths. This is confirmed by the confusion matrix in Figure~\ref{fig:conf_mat}, which shows that the majority of estimates are within one atom of the ground-truth depth.

\subsection{Calibration}
\begin{figure}[t]
    \centering
  \subfloat{%
       \includegraphics[width=0.65\linewidth]{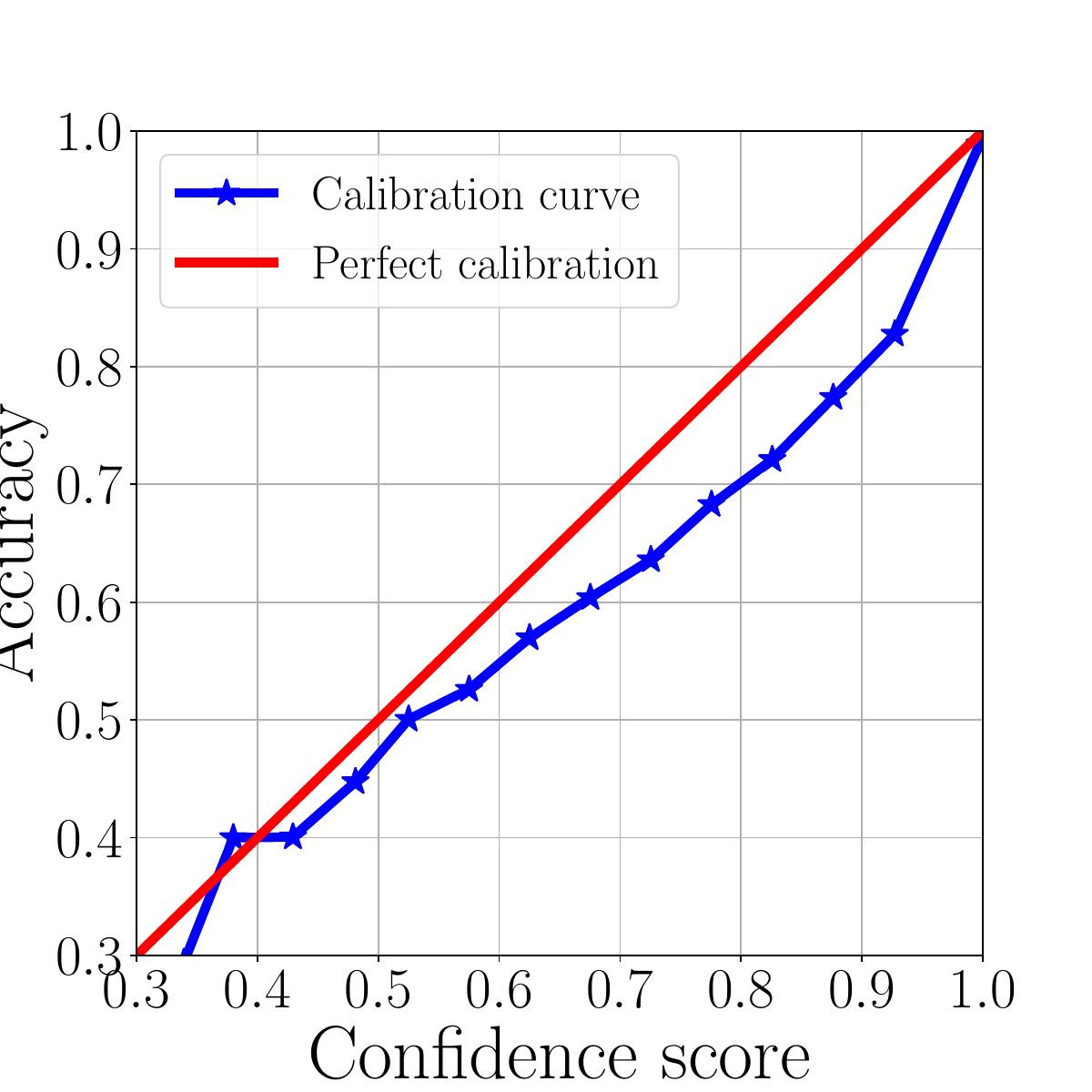}}
        \caption{\textbf{Calibration of model output.} We compare the predicted confidence score, defined in equation \eqref{eq:confidence}, of the proposed model (with 64 base channels) over the test set of simulated TEM images to the model accuracy, aggregated over bins of pixels with similar confidence scores. The confidence score is reasonably calibrated, the calibration curve is close to a diagonal line that would indicate perfect calibration. } 
        \label{fig:calibration}
\end{figure}

 Figure \ref{fig:calibration} evaluates to what extent the confidence score, defined in equation \eqref{eq:confidence}, is well calibrated~\cite{guo2017calibration,liu2022deep}, in the sense that it correctly quantifies the uncertainty in the model estimate. To assess calibration we compare the predicted confidence score (computed over the test set) to the model accuracy, aggregated over bins of pixels with similar confidence scores. The confidence score is well calibrated, but tends to underestimate the uncertainty. For example, the accuracy for pixels with calibration scores close to 0.8 is around 0.7.
 
\begin{figure}[h!]
\captionsetup[subfigure]{labelformat=empty,skip=0.3\baselineskip}
\begin{minipage}[c]{.3\columnwidth}
\centering
\vspace{-2cm} (a) \vspace{1.5cm}\\
Output\\
\subfloat[]{\includegraphics[width=\columnwidth]{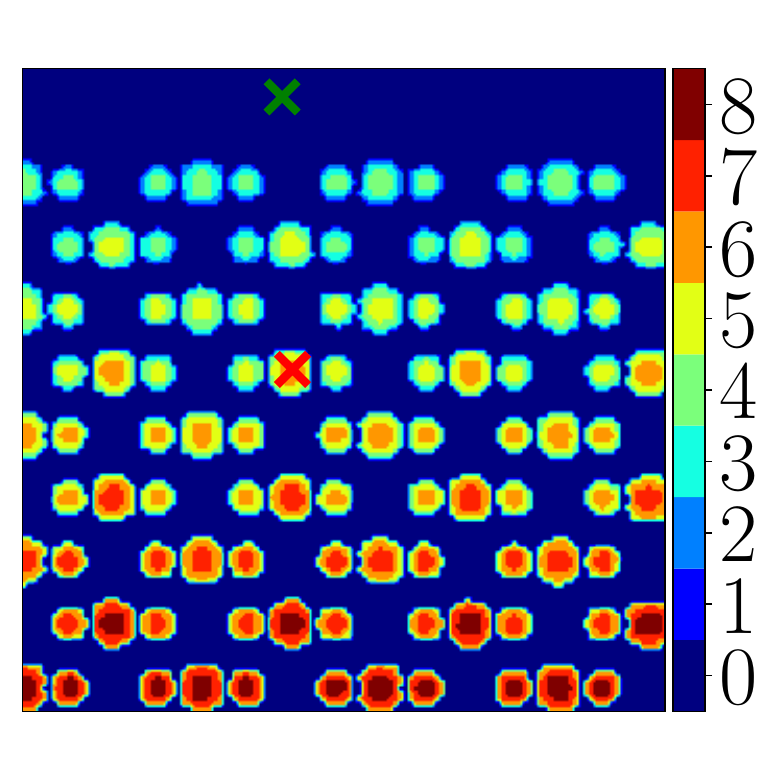}}%
\end{minipage}%
\hspace{.05\linewidth}%
\begin{minipage}{.65\columnwidth}
\centering
Gradient w.r.t. red-marker pixel\\\vspace{-0.2cm}
%% 0.461 = 0.3/0.65
\subfloat[Depth 5]{\includegraphics[width=0.49\columnwidth]{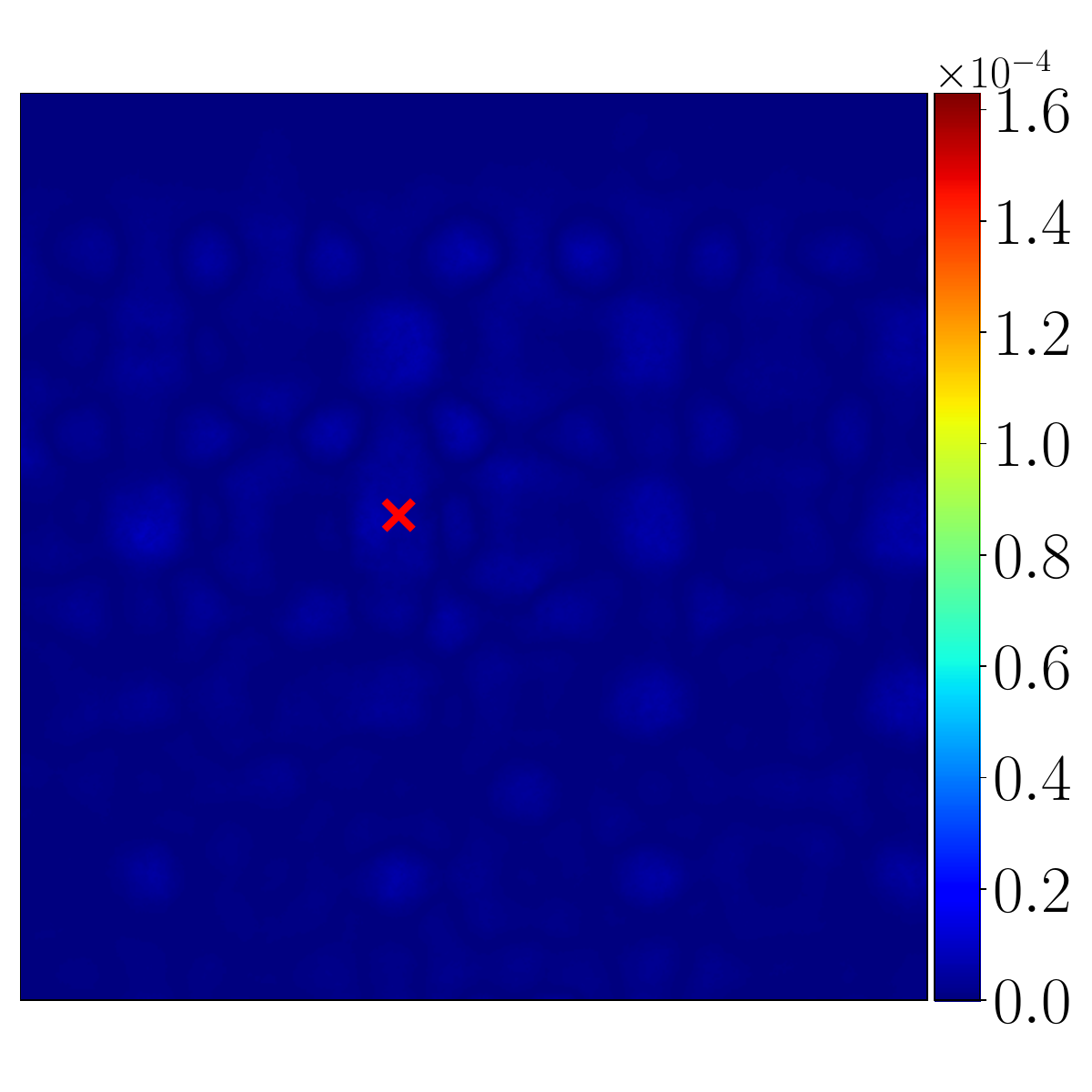}}%
\hspace{\fill} 
\subfloat[Depth 6]{\includegraphics[width=0.49\columnwidth]{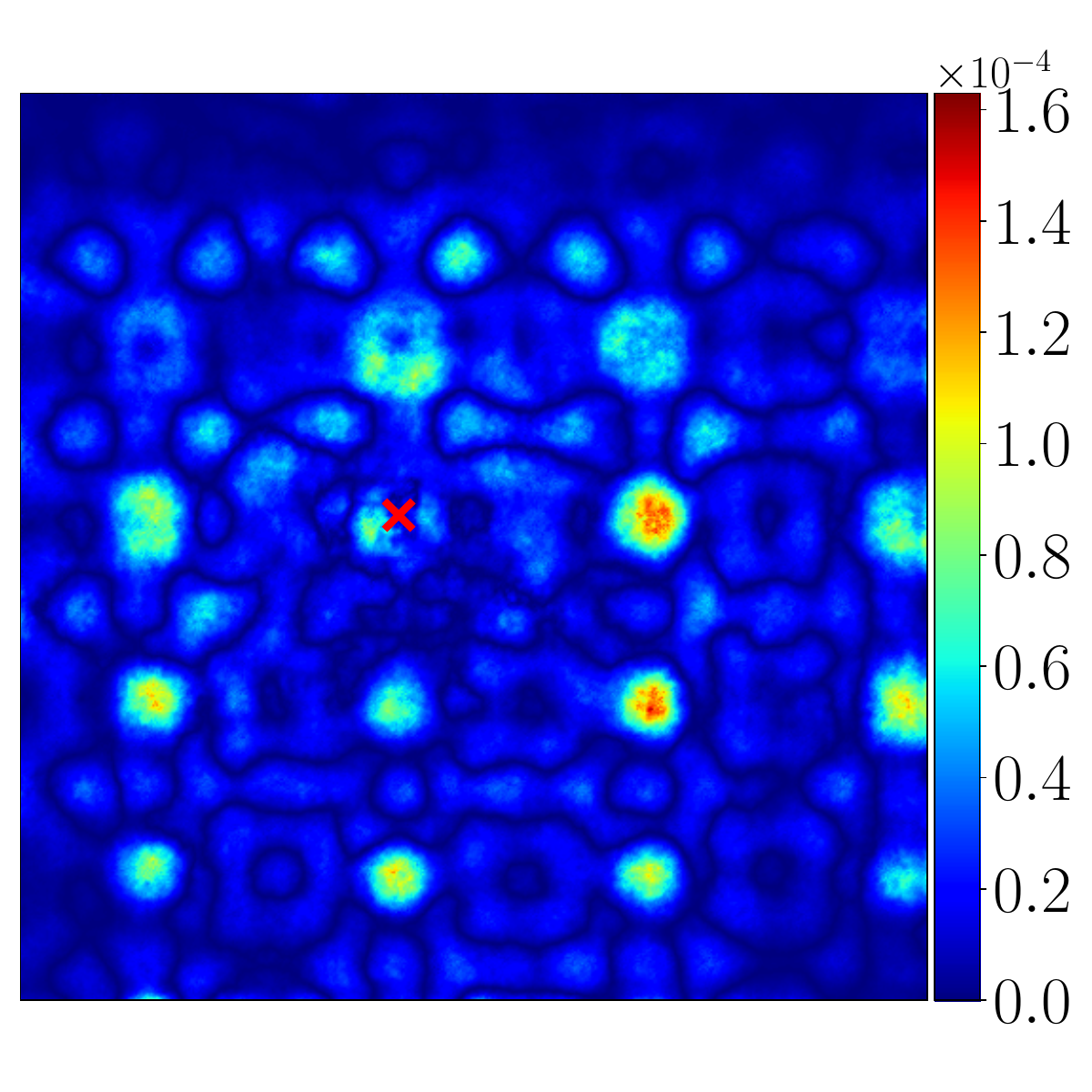}}\vspace{0.4cm}\\
Gradient w.r.t. green-marker pixel\\
\vspace*{-0.9em}
\subfloat[Depth 0]{\includegraphics[width=0.49\columnwidth]{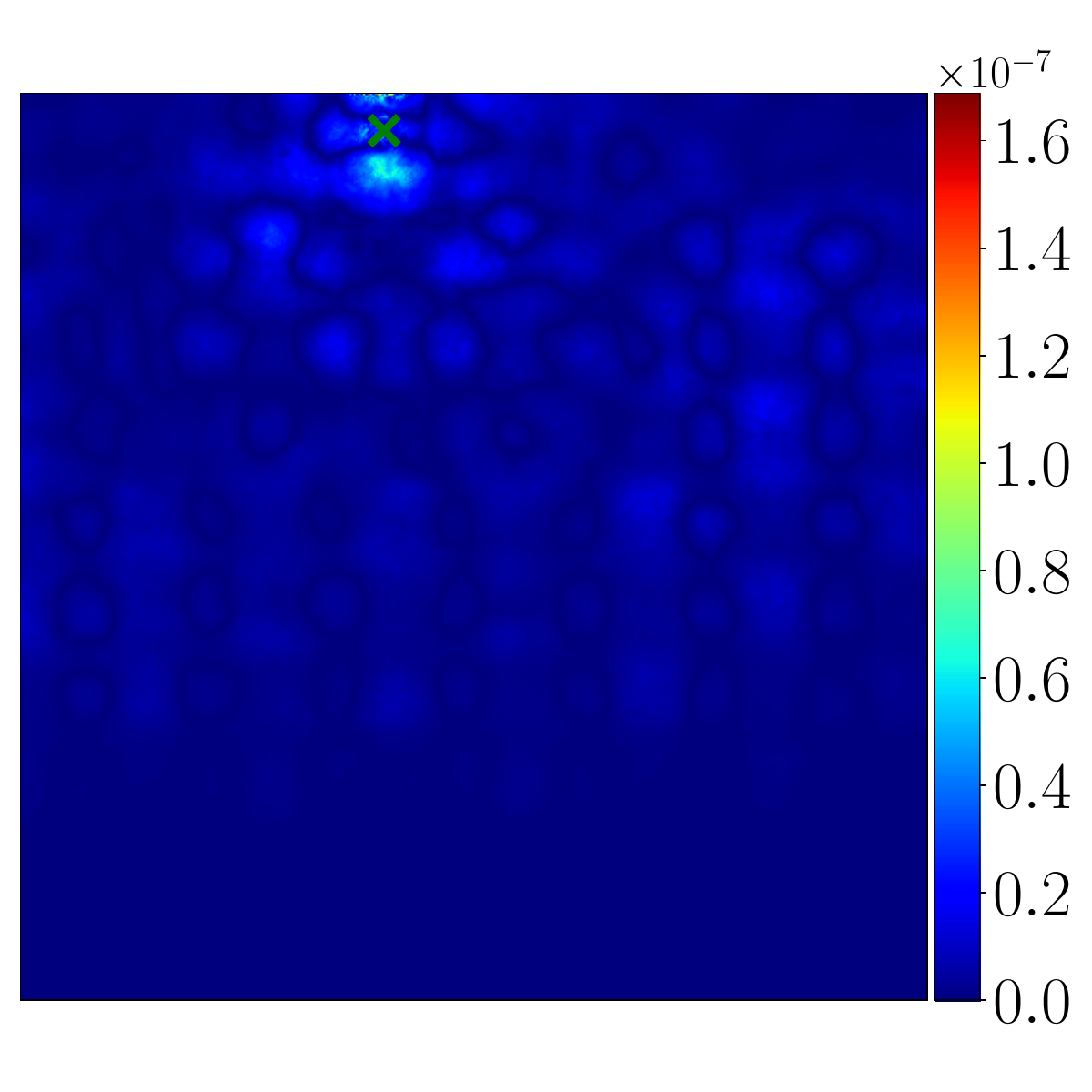}}%
\hspace{\fill}
\subfloat[Depth 1]{\includegraphics[width=0.49\columnwidth]{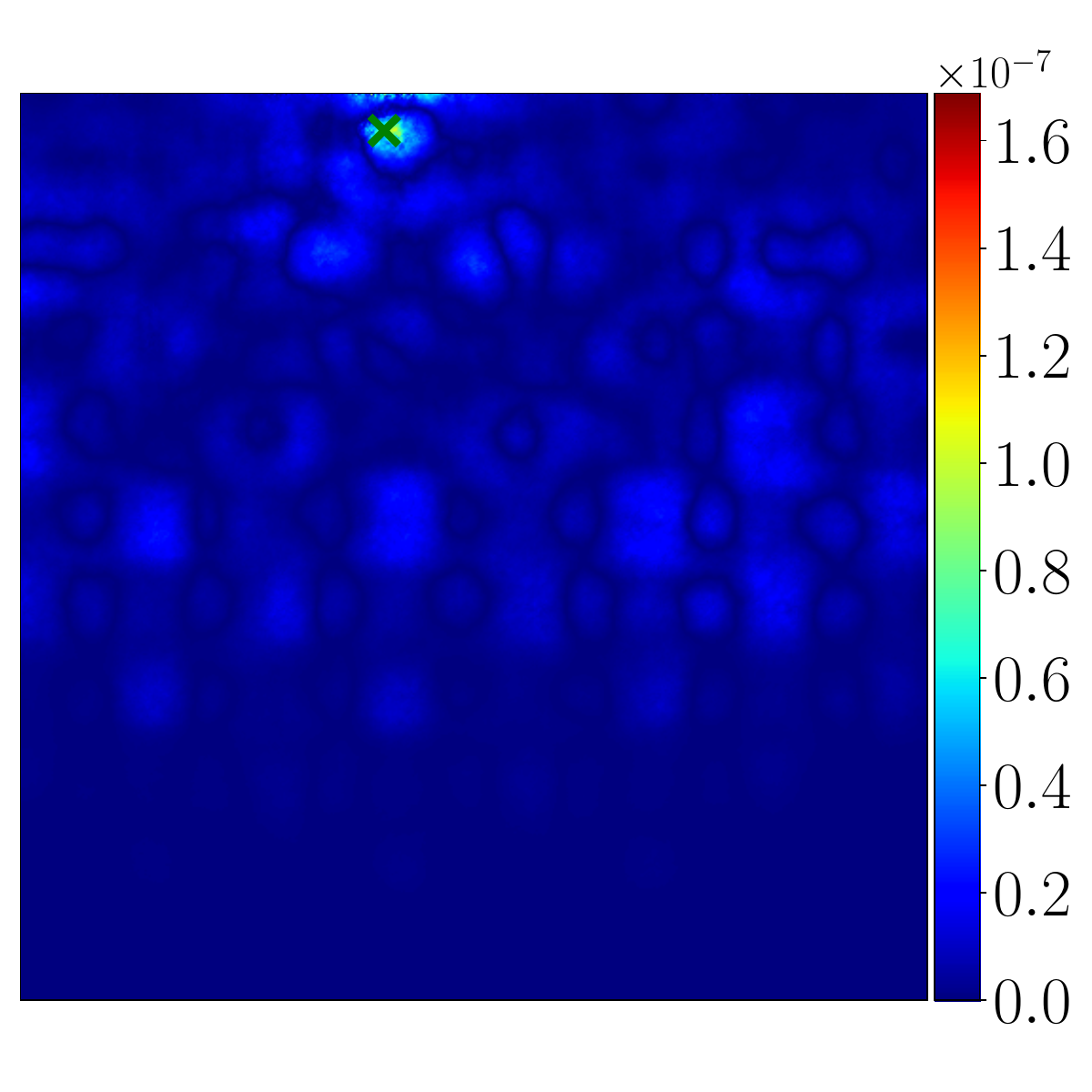}}
\end{minipage} \vspace{0.25cm}\\
\centering 
(b) Dependence with signal to noise ratio\vspace{0.2cm}\\
\begin{tabular}{@{}c@{}c@{}c@{}}
\includegraphics[width=0.33\columnwidth]{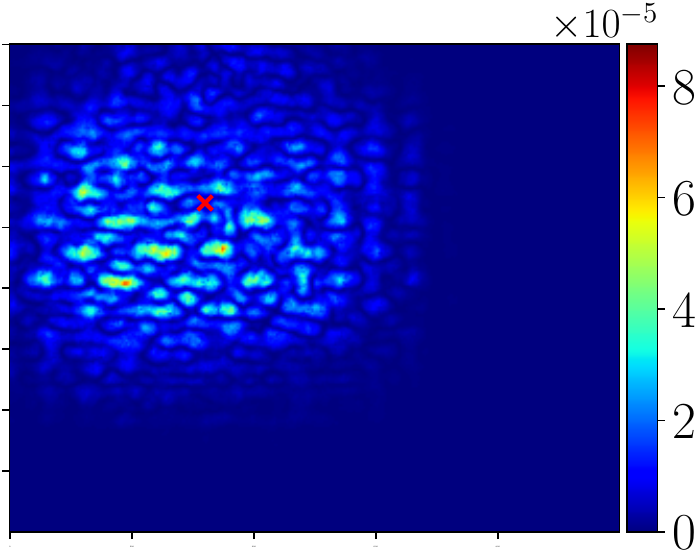} &\includegraphics[width=0.33\columnwidth]{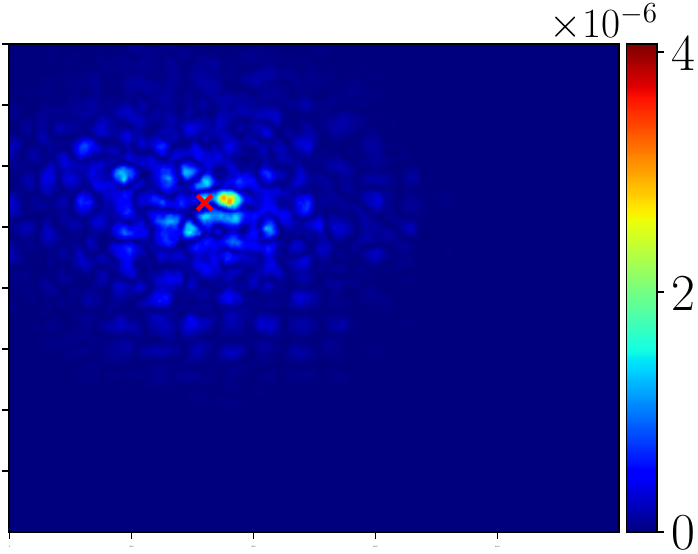}&
\includegraphics[width=0.33\columnwidth]{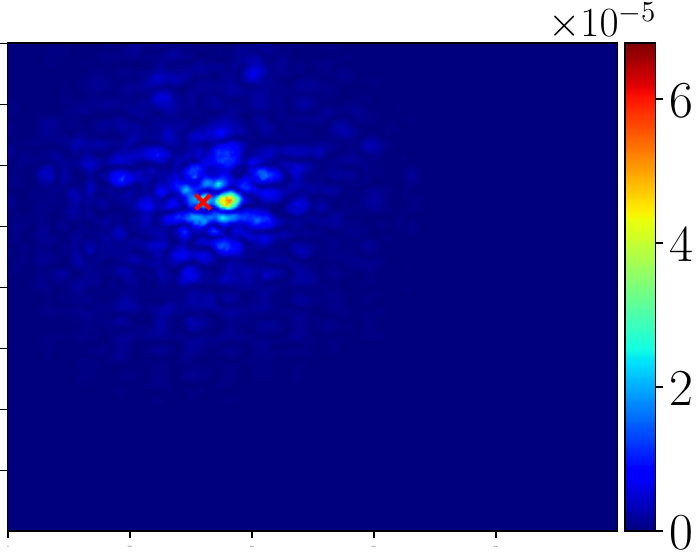}\\
$\lambda$ = 0.1&$\lambda$ = 1&$\lambda$ = 1.5
\end{tabular}
\caption{\textbf{Gradient visualization.} The magnitude of the gradient of the neural-network output with respect to its input quantifies the sensitivity of the estimate to variations in the input pixels. (a) We consider two pixels indicated by a red and a green marker in the left image, which depicts the model output (depth estimate) for a simulated TEM image. The red marker is on an atomic column with depth 6. The magnitude of the gradient of that output pixel with respect to the input is highest for depths 5 and 6 (top right), and follows a periodic structure, which suggests that the model exploits long-range context. The green marker is in the background. The corresponding gradient magnitude is highest for depths 0 and 1 (bottom right) and is much more localized, indicating less sensitivity to long-range context. (b) Gradient magnitude  of the pixel indicated by the red marker when the input to the model is the same clean TEM image corrupted using different values of the noise parameter $\lambda$. The lower the SNR, the more spread out the gradient is, indicating that the model uses longer-range context to obtain the depth estimate.}
% \caption{\textbf{Simulated Gradient map}: 
% We present an example of an output and the dominant channel gradients for a pixel located on the particle (red marker), and in the vacuum (green marker).
% On the particle, the pixel's value is 6, and we can see that indeed channels 5,6 are the most active. The gradient is not local  and follows a periodical structure, which suggest that in the presence of noise, as expected, the network requires long range context to classify the pixel correctly. In contrast, in the vacuum, we can see that only channels 0,1 are active. The gradient is much more localized on the vacuum, so the network correctly identifies the boundary of the particle. }
        \label{fig:grad_sim}
\end{figure}
\subsection{Gradient analysis}
\label{sec:gradient_analysis}
A popular approach to interpret the functions learned by deep neural networks is to compute the gradient of the output with respect to their input~\cite{simonyan2013deep,Mohan2020Robust,mohan2025learn}. As described in Section~\ref{sec:deep_learning} the SegDepth network utilizes a UNet that receives a TEM image as an input and outputs 11 probability maps (corresponding to the possible depth values) with the same dimensions as the input image. For each output pixel $i$, each possible depth value $d$ and an input image $y$ we can compute the gradient of the estimated probability with respect to $y$,
\begin{align}
g(i,d) := \nabla_{y} F_{\theta}(y)_i[d]. 
\end{align}
Note that here the gradient is \emph{not} computed with respect to the network parameters $\theta$, as during training, but rather with respect to the input image. Intuitively, the magnitude of $g(i,d)$ quantifies the sensitivity of the probability estimate for depth $d$ at pixel $i$ to variations in the input pixels. 

 \begin{figure}[t]
 \centering
 \begin{tabular}{c}
   \includegraphics[width=0.95\columnwidth]{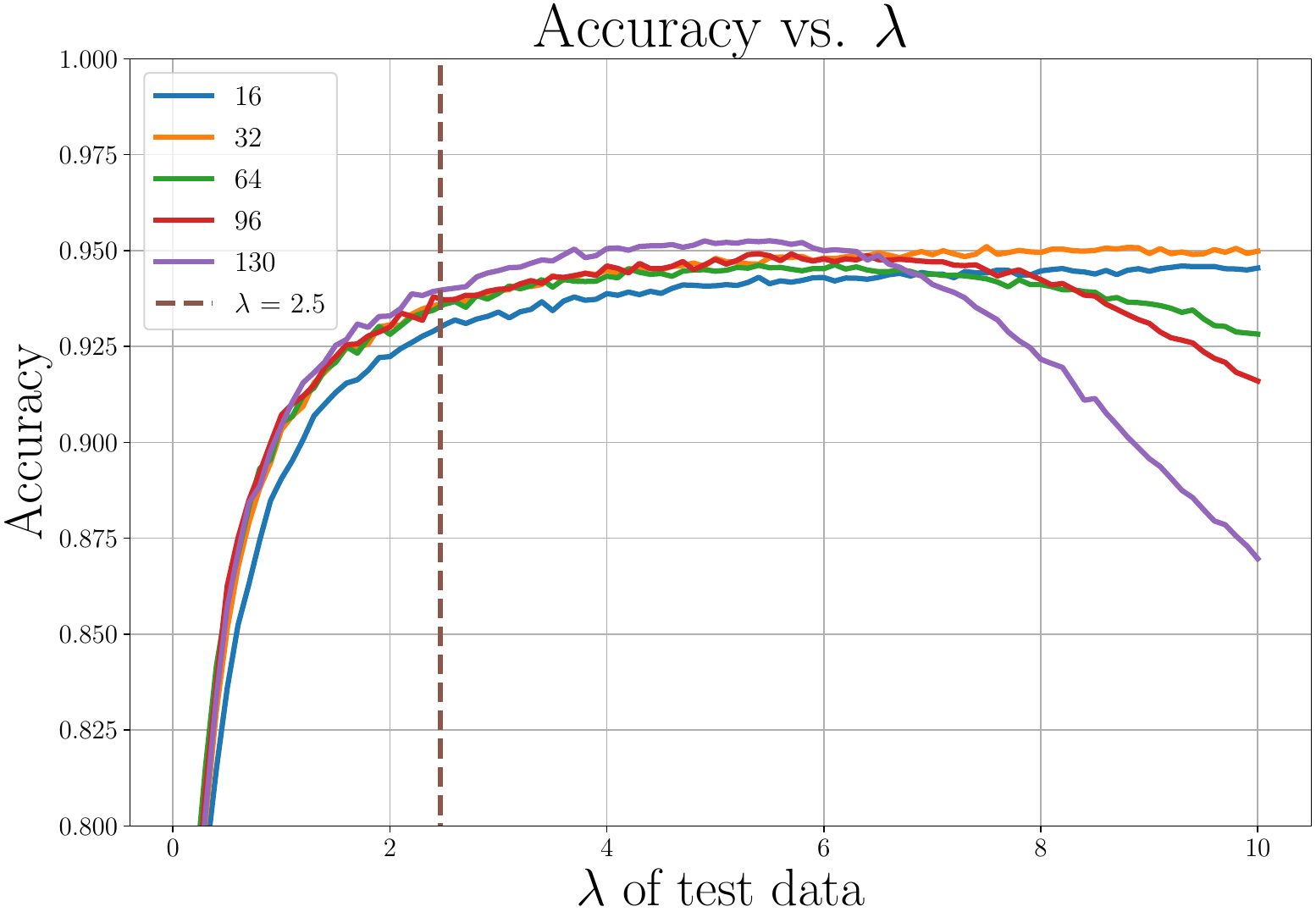} \\\small{$(a)$}\\\includegraphics[width=0.95\columnwidth]{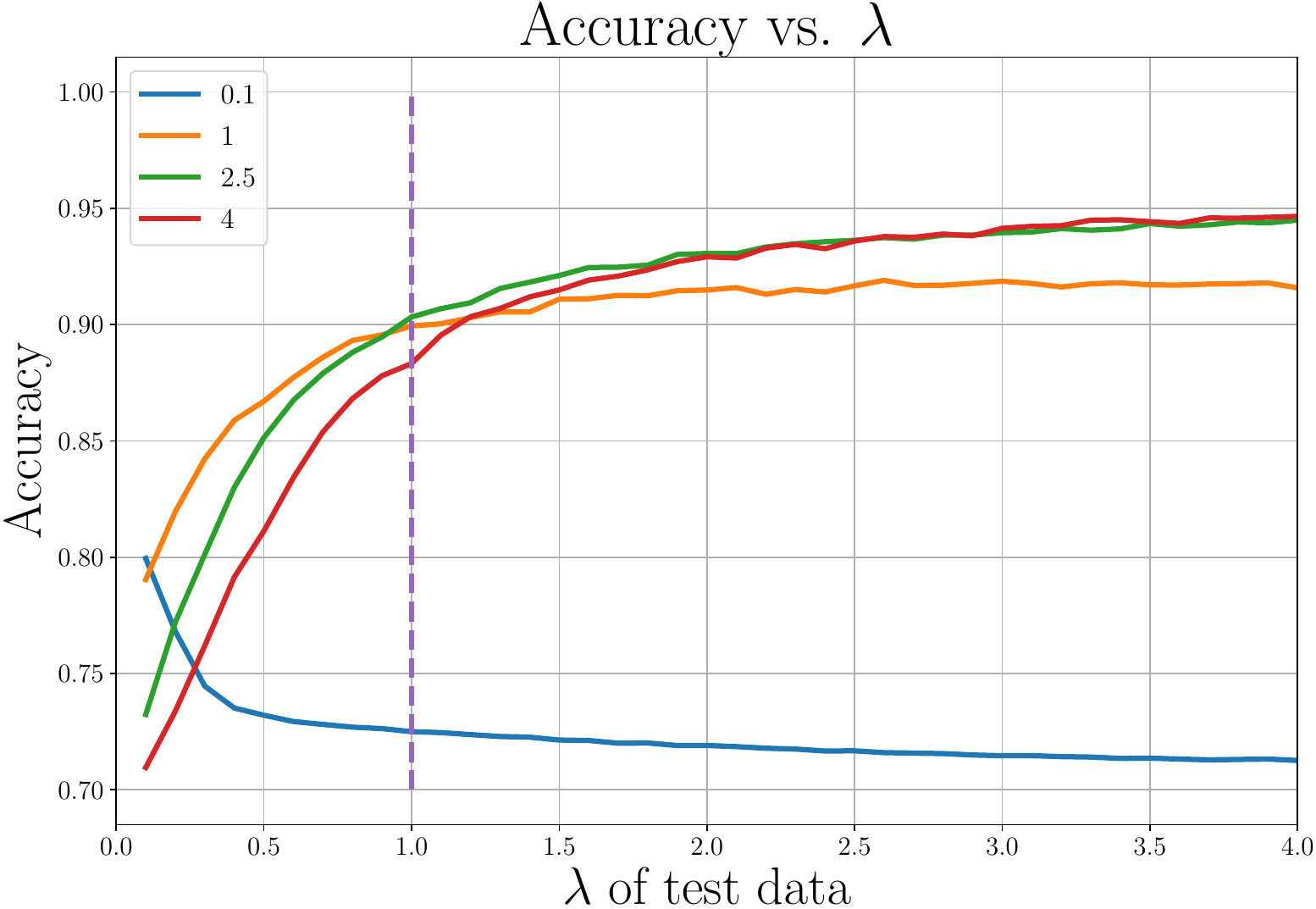}\\
   \small{$(b)$}
 \end{tabular}
 \caption{\textbf{Generalization to different noise levels.} (a) We compare the generalization ability of models with different number of parameters when they are applied to data with different values of the noise parameter $\lambda$ (horizontal axis), which determines the SNR as justified by \eqref{eq:lambda}. All models were trained with noisy data where $\lambda$ is equal to 2.5. Models with fewer parameters generalize better to data with higher SNR. (b) We compare the generalization ability of models with 16 base channels trained using noisy data with different values of the noise parameter $\lambda$ (indicated by the legend) applied to test sets generated with different values of $\lambda$ (horizontal axis). Models tend to perform better close to the training noise level, generalizing well across values of $\lambda$ with the same order of magnitude. However, generalization no longer occurs when $\lambda$ is very different at inference and in training, as evinced by the poor generalization of the model trained with $\lambda := 0.1$.}
        % \caption{  $(a)$ \textbf{SNR generalization vs. model complexity}: We compare the average accuracy of networks trained on a particular SNR to lower and higher values. All networks were trained at an SNR of 2.466, and differ based on the number of base channels in the UNet architecture. We see that there is a trade off between robustness and accuracy for a particular SNR. Models will fewer parameters will generalize well on images with higher SNR, and less so for lower SNR, while higher complexity models will not generalize as well to higher SNR. $(b)$ \textbf{SNR generalization vs. training point}: We compare the average accuracy of networks trained on different SNRs to lower and higher values.  Training on extremely low SNR provides marginally better results than other models but doesn't generalize. Training on high SNR achieves higher accuracy but deteriorates faster as the SNR decreases. All networks have 16 base channels.  }
        \label{fig:generalization_noise}
\end{figure}

In Figure~\ref{fig:grad_sim}(a) we apply this gradient analysis to two pixels, indicated by a red and a green marker, in a simulated TEM image. The red-marker pixel belongs to an atomic column with depth 6 and the green-marker belongs to the background. The magnitude of the gradient of both pixels with respect to the input is highest for depths close to the ground-truth (5 and 6 for the red marker, 0 and 1 for the green marker). For the background pixel, the gradient is very localized, indicating that the output is mostly sensitive to a small neighborhood around the pixel. In contrast, for the atomic-column pixel, the gradient magnitude is large at other atomic columns, suggesting that the model is able to exploit the periodic  atomic structure of the nanoparticle. Figure~\ref{fig:grad_sim}(b) illustrates this phenomenon with another example, where we vary the SNR, keeping the underlying TEM image the same. At low SNR, the gradient is spread out, indicating that the model uses other atomic columns to obtain the depth estimate. As the SNR increases, the gradient becomes more localized, possibly because it is no longer necessary to rely on additional context.

\subsection{Generalization to different noise levels}
 An important consideration, when evaluating neural networks designed to process real noisy data, is to what extent they generalize to different noise levels. Figure~\ref{fig:generalization_noise}(a) reports the performance of several models with different number of parameters when we apply them to data with an SNR that is different from the training SNR. The models are able to generalize over a range of noise levels within an order of magnitude. Larger models tend to be less robust to changes in SNR; their performance deteriorates faster as the SNR increases beyond the training SNR. Figure~\ref{fig:generalization_noise}(b) compares models that share the same architecture, but were trained at different noise levels, reporting their performance over a range of test noise levels. The models tend to perform better close to their training noise level and generalize robustly. However,  performance can deteriorate dramatically at higher SNR if the training SNR is too low (as evinced by the model trained with noise parameter $\lambda := 0.1$). 

 \begin{table*}[t]
     \caption{Performance comparison between the proposed framework (Segmentation) and an alternative approach, where the data are first denoised and then segmented. Direct segmentation clearly outperforms two variants of this approach where the denoising and segmentation models are trained sequentially (Denoising $\rightarrow$ Segmentation) and jointly (Denoising $+$ Segmentation). All models have 64 base channels.}
    \centering
    \begin{tabular}{cccccc}
    \toprule
        Method & Pixelwise&Center &Real Atom&Hallucinated Atom\\
        &Accuracy&Accuracy&Detection Rate&Rate ($\times 10^{-2}$)\\\midrule
        Segmentation & \textbf{0.93}&\textbf{0.847}&\textbf{0.956}&\textbf{2.33}\\
        Denoising $\rightarrow$ Segmentation&0.867&0.765&0.946&12.35\\
        Denoising $+$ Segmentation &0.862&0.704&0.936&17.38\\\bottomrule
    \end{tabular}
    \label{tab:seg_denois}
\end{table*}
 
\subsection{Denoising and segmentation}
In the proposed SegDepth framework, a neural network is trained to perform segmentation directly from noisy data. However, a possible alternative could be to first denoise the TEM images using a separate neural network and then perform segmentation using the denoised images. We consider two variants of this approach. 

In the first variant, the denoising network is trained independently on noisy-clean pairs of TEM images in a supervised manner, minimizing the mean squared error~\cite{mohan2022deep,vincent2021developing}. The segmentation network is then trained using the denoised TEM images and the corresponding depth segmentation maps, as described in Section~\ref{sec:deep_learning}. 

In the second variant, the denoising and segmentation networks are trained jointly. Each noisy TEM image is fed into the denoising network. The output is compared to a clean image to optimize the parameters of the denoising network. Then this output is fed into the segmentation network and the corresponding depth segmentation map is used to optimize the parameters of the segmentation network.

Table~\ref{tab:seg_denois} shows that SegDepth outperforms both variants in all metrics. In particular, its hallucinated atom rate is much lower. The first variant which trains the two networks sequentially is better than the second, indicating that it may be challenging to jointly optimize the denoising and segmentation losses.

% \begin{figure}[htbp!] 
% \begin{tabular}{c c c}
% \includegraphics[width=0.333\columnwidth]{plots_Dec22/grad_real_particle_diff_snr0_v2.pdf}&\includegraphics[width=0.333\columnwidth]{plots_Dec22/grad_real_particle_diff_snr1_v2.pdf}&
% \includegraphics[width=0.333\columnwidth]{plots_Dec22/grad_real_particle_diff_snr2_v2.pdf}\\
% SNR=0.1&SNR=1&SNR=1.5
% \end{tabular}
%         \caption{\textbf{SNR generalization vs. training point}: We compare the average accuracy of networks trained on different SNRs to lower and higher values.  Training on extremely low SNR provides marginally better results than other models but doesn't generalize. Training on high SNR achieves higher accuracy but deteriorates faster as the SNR decreases.  } 
%         \label{fig:grad_snr}
% \end{figure}
\section{Real experimental data}
 \label{sec:real_dataset}
 We applied the proposed framework SegDepth to a time-resolved TEM dataset consisting of a series of images of CeO$_2$ catalyst (110) surfaces viewed along the [110] zone axis direction. The images were acquired using an aberration-corrected FEI Titan  environmental transmission electron microscope (ETEM), operated at 300 kV and coupled with a Gatan K3 IS direct electron detector camera. The detector was operated in electron counting mode with a time resolution of 0.13 sec/frame and an incident electron dose rate of 5,000 e$^-$/{\AA}$^2$/s. The  lens system of the microscope was tuned to achieve a highly coherent parallel beam configuration with minimal low-order aberrations (e.g., astigmatism, coma), and a third-order spherical aberration coefficient of approximately -9 $\mu$m.

\begin{figure}[t]
%\resizebox{\columnwidth}{!}{
\begin{tabular}{@{}c@{}c@{}c@{}}%{ccc}
    \includegraphics[width=0.34\linewidth]{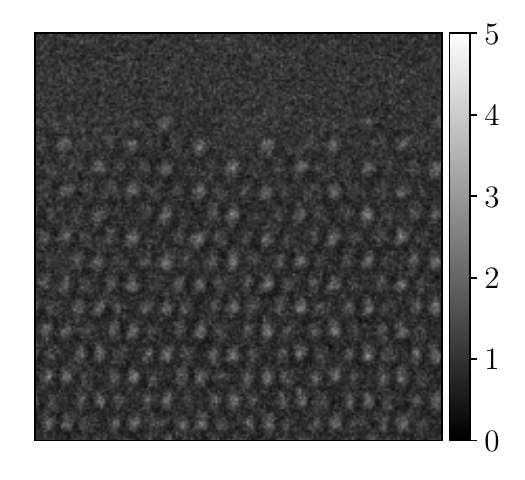} &  \includegraphics[width=0.34\linewidth]{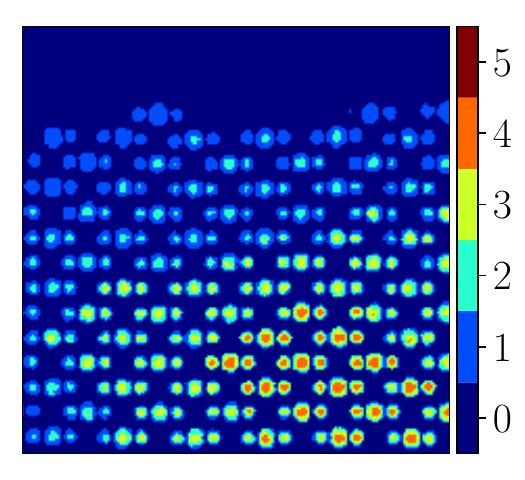}&\includegraphics[width=0.34\linewidth]{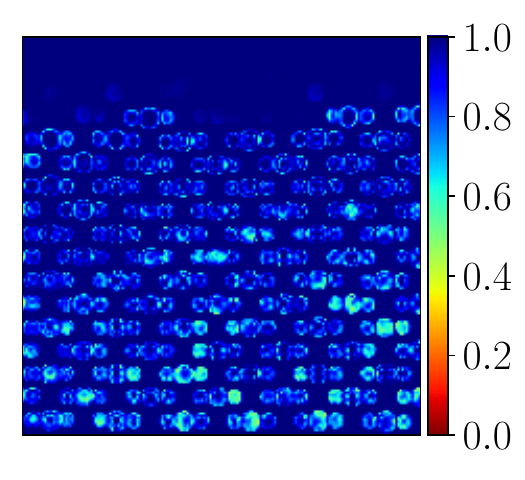}\\\small{$(a)$ Real data}&\small{$(b)$ Depth estimate}&\small{$(c)$ Confidence score}
\end{tabular}
%}
        \caption{\textbf{Application to real data.} 
        (a) Real TEM image of a  CeO$_2$ nanoparticle viewed along the [110] zone axis direction. (b) Depth estimate produced by the proposed framework SegDepth. The atomic column depth decreases towards the nanoparticle surface, which is expected. (c) Confidence score of the proposed model at each pixel, computed as defined in \eqref{eq:confidence}. The confidence is lower in regions where the depth estimate changes value, reflecting a greater uncertainty.}
        %The network applied to a real data sample. $(a)$ The input data with an estimated SNR of 1. $(b)$ The network's output. We see that the atomic column depth decreases as one reaches the surface which is the expected result $(c)$ Pixel valued confidence score. The confidence score is lower than the simulated image. That is expected since the SNR is lower} 
\label{fig:real_example}
\end{figure}

\begin{figure}[t]
\begin{comment}
\captionsetup[subfigure]{labelformat=empty,skip=0.3\baselineskip}
% \begin{minipage}[c]{.25\columnwidth}
% \vspace*{-7.5cm}
% \centering
% Output\\
% \subfloat[]{\includegraphics[width=\linewidth]{plots_Dec22/real_particle_output_with arrows.pdf}}%
% \end{minipage}%
% \hspace{.05\linewidth}%
\centering
\begin{minipage}[b]{0.65\columnwidth}
\centering
%% 0.461 = 0.3/0.65
\subfloat[Particle \# 0]{\includegraphics[width=0.49\columnwidth]{plots_Dec22/grad_realdata_particle_chan0_v2.pdf}}%
\hspace{\fill} 
\subfloat[Particle \# 1]{\includegraphics[width=0.49\columnwidth]{plots_Dec22/grad_realdata_particle_chan1_v2.pdf}}
\\
\vspace*{-0.9em}
\subfloat[Vacuum \# 0]{\includegraphics[width=0.49\columnwidth]{plots_Dec22/grad_realdata_vacuum_chan0_v3.pdf}}%
\hspace{\fill}
\subfloat[Vacuum \# 1]{\includegraphics[width=0.49\columnwidth]{plots_Dec22/grad_realdata_vacuum_chan1_v2.pdf}}
\end{minipage}
\centering
%\begin{tabular}{@{}>{c@{}c@{}c@{}}
Red marker (nanoparticle boundary)\vspace{0.1cm}\\
\begin{tabular}{@{}c@{}c@{}}
\includegraphics[width=0.49\columnwidth]{plots_Dec22/grad_realdata_particle_chan0_v2.pdf}   & \includegraphics[width=0.49\columnwidth]{plots_Dec22/grad_realdata_particle_chan1_v2.pdf} \\
Depth 0 & Depth 1 
\end{tabular} \vspace{0.3cm}\\
Green marker (vacuum)\vspace{0.1cm}\\
\begin{tabular}{@{}c@{}c@{}}\includegraphics[width=0.49\columnwidth]{plots_Dec22/grad_realdata_vacuum_chan0_v3.pdf}  & \includegraphics[width=0.49\columnwidth]{plots_Dec22/grad_realdata_vacuum_chan1_v2.pdf}\\
Depth 0 & Depth 1 
\end{tabular}
\end{comment}
\centering
\begin{minipage}[c]{\columnwidth}
\centering
Depth estimate \vspace{-0.1cm}\\
\includegraphics[width=0.4\linewidth]{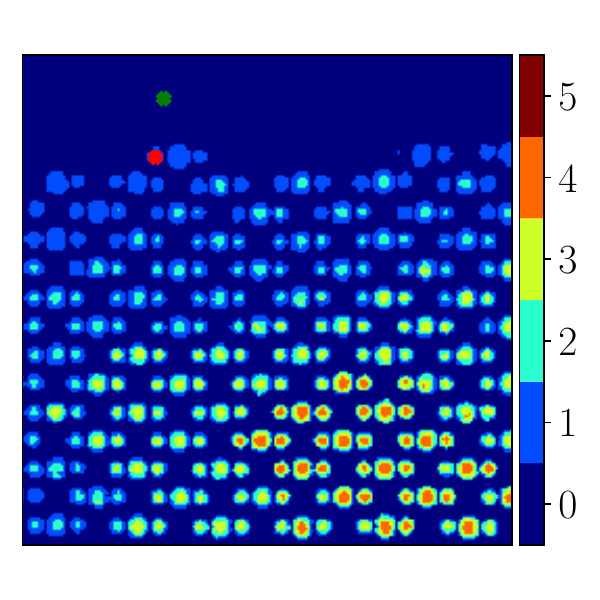}%
\end{minipage}
\begin{tabular}{@{}c@{}c@{}}
Gradient w.r.t. red-marker  & Gradient w.r.t. green-marker\\
pixel (boundary) & pixel (vacuum)\\
\includegraphics[width=0.49\columnwidth]{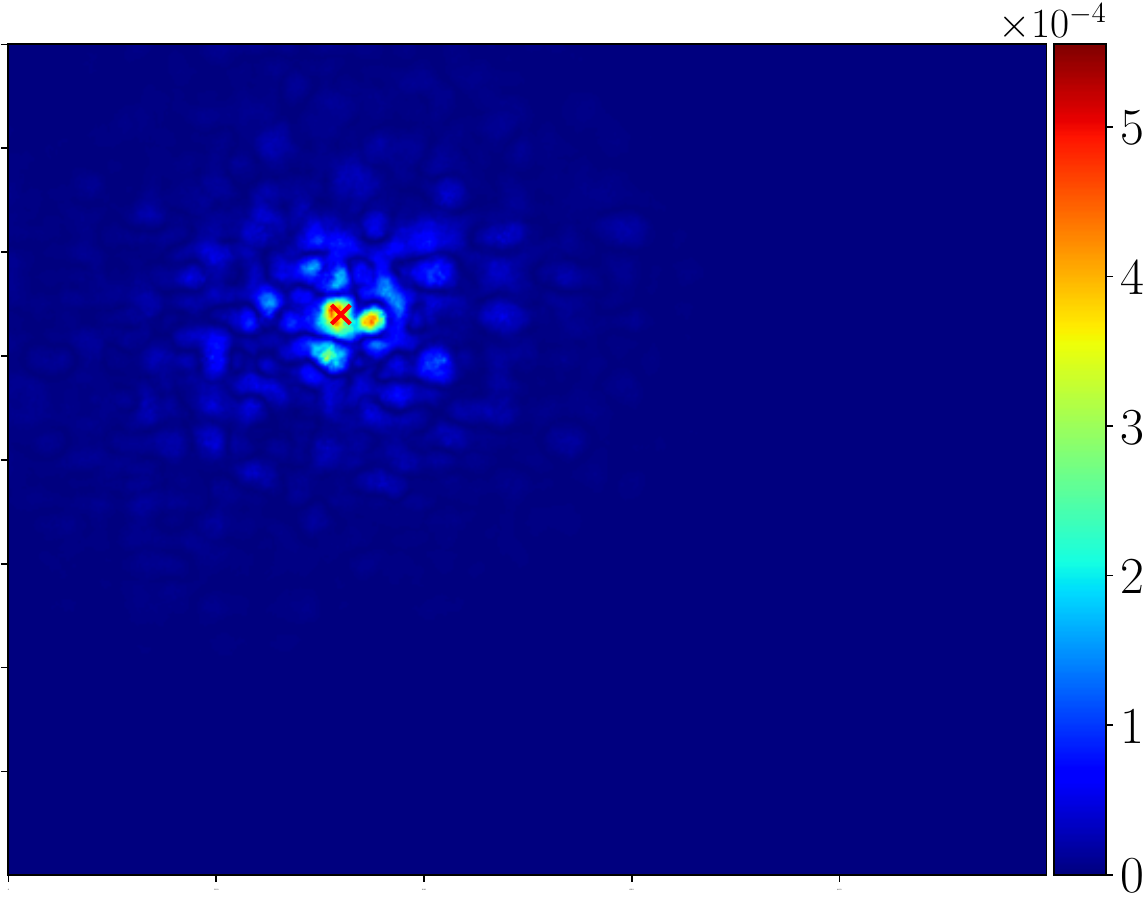}   & \includegraphics[width=0.49\columnwidth]{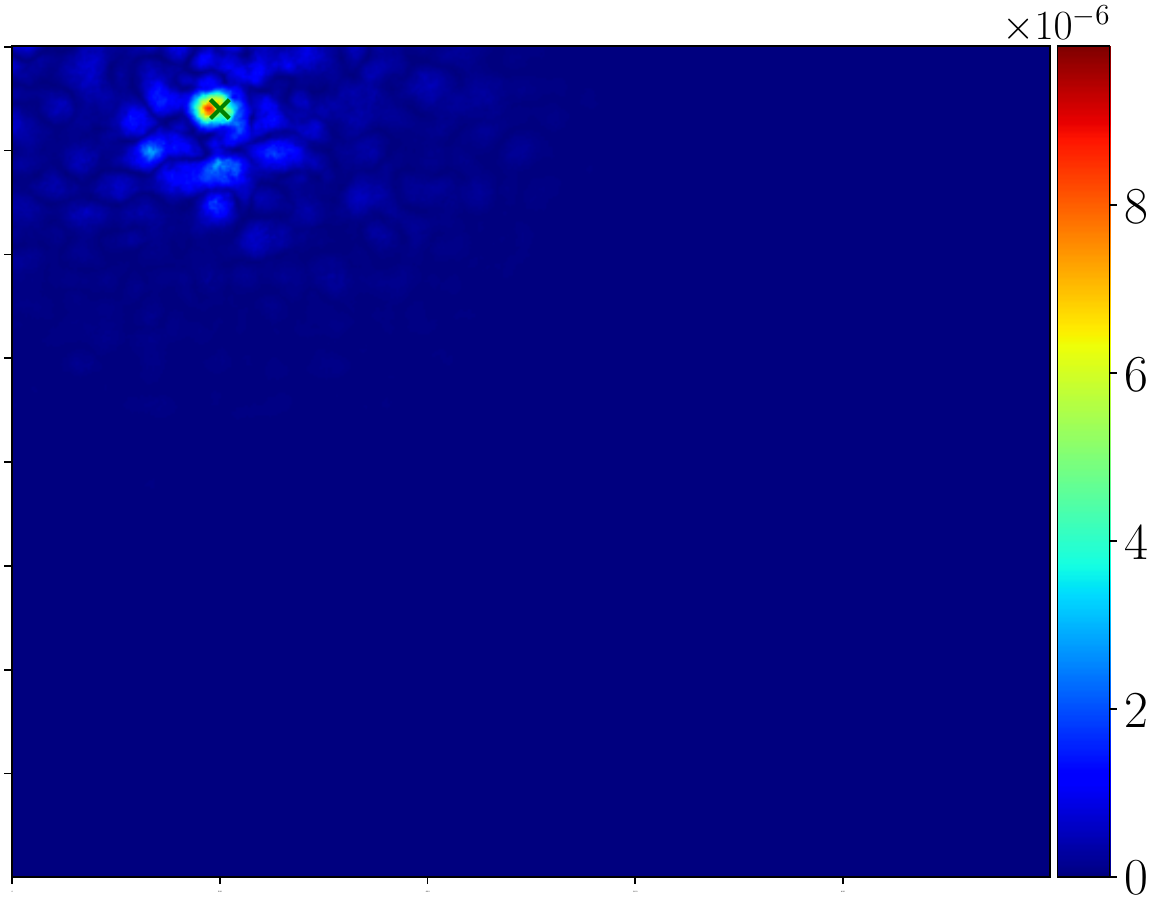} \\
Depth 1 & Depth 0 
\end{tabular} 
\caption{\textbf{Gradient analysis of real data.} We apply the gradient analysis in Section~\ref{sec:gradient_analysis} to two pixels, indicated by a red and a green marker, in the TEM image from Figure~\ref{fig:real_example}. The red marker is at the boundary of the nanoparticle, whereas the green marker is in the vacuum. The images show the depths for which the gradient magnitude is highest: 1 for the red marker and 0 for the green marker. The gradient is mostly localized in a neighborhood of the pixels,  indicating low sensitivity to long-range context at the two locations.} 
        \label{fig:grad_real}
\end{figure}

Although real datasets lack ground truth for evaluation, qualitative inspection of the results was promising. As illustrated in Figure~\ref{fig:real_example}, the depth estimate correctly identifies the positions of the atomic columns. The depth values decrease in the direction of the vacuum, which is consistent with the change in contrast between Ce and O atoms. The pixelwise confidence score, computed following \eqref{eq:confidence}, is mostly high, with some lower values indicating uncertainty in transition regions, where the depth estimate changes. In Figure~\ref{fig:grad_real} we apply the gradient analysis in Section~\ref{sec:gradient_analysis} at two locations: the boundary of the nanoparticle and the surrounding vacuum. The gradient is mostly localized in a small neighborhood, indicating low sensitivity to long-range context.

% \subsubsection{Tracking dynamical changes using a segmentation network}
% The real data consists of 75 centered frames, so we can also track the dynamics of the three dimensional structure of the sample. In Figure~\ref{fig:output_real} we focus on two pixels, one in the vacuum and on an oxygen atom close to the surface, and follow their dynamics.

\begin{figure}[t]
% \captionsetup[subfigure]{labelformat=empty,skip=0.3\baselineskip}
% \begin{minipage}[c]{\columnwidth}
% \centering
% Depth estimate\\
% \vspace*{-1.5em}
% \subfloat[]{\includegraphics[width=0.4\linewidth]{plots_Dec22/real_particle_output_with arrows.pdf}}%
% \end{minipage}%
%\begin{minipage}{0.69\columnwidth}\\
%\vspace*{-0.8em}
%\hspace{1em}
\begin{tabular}{@{}c@{}c@{}}
  Red marker & Green marker\\
   (nanoparticle boundary)  & (vacuum)\\
\includegraphics[width=0.5\linewidth]{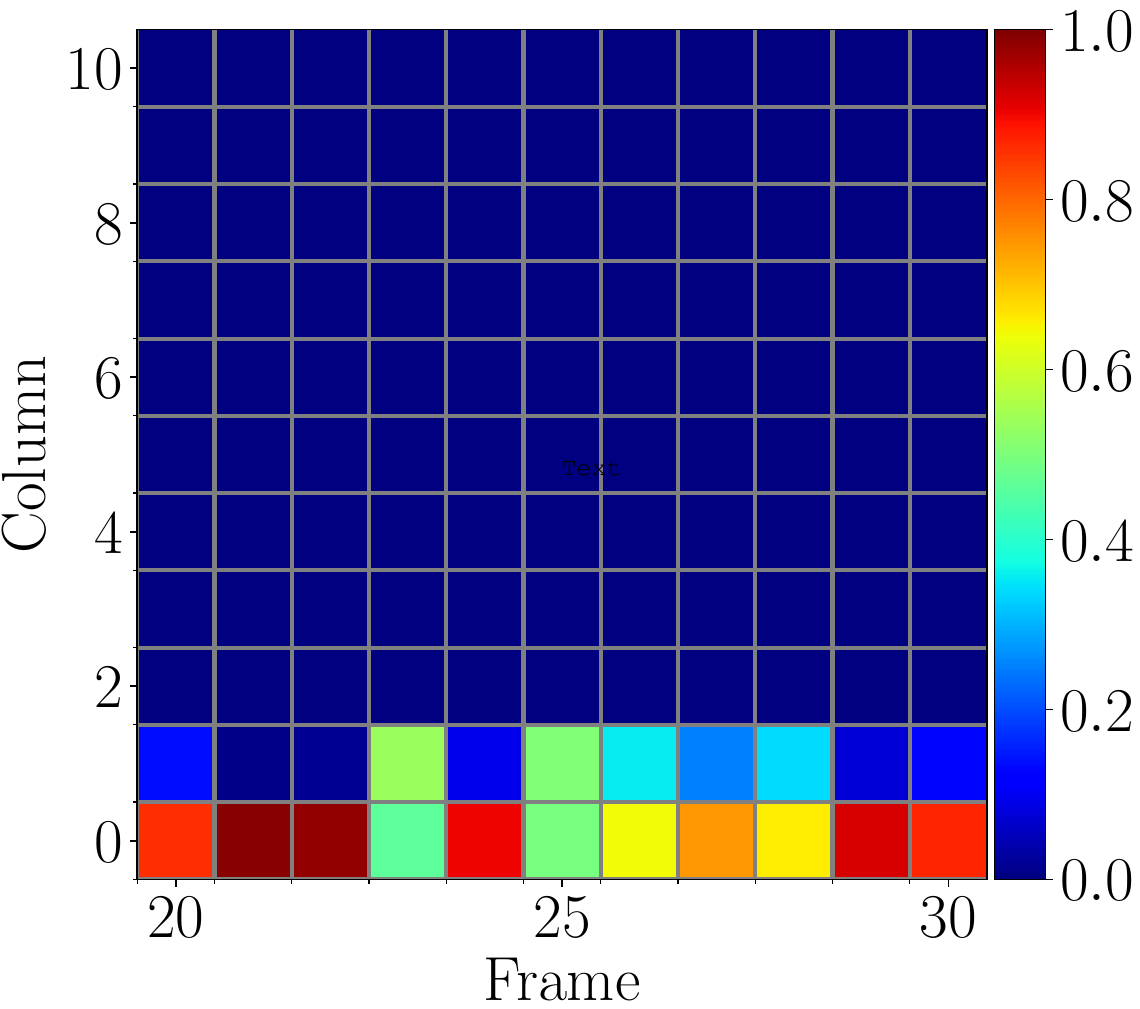}  &
\includegraphics[width=0.5\linewidth]{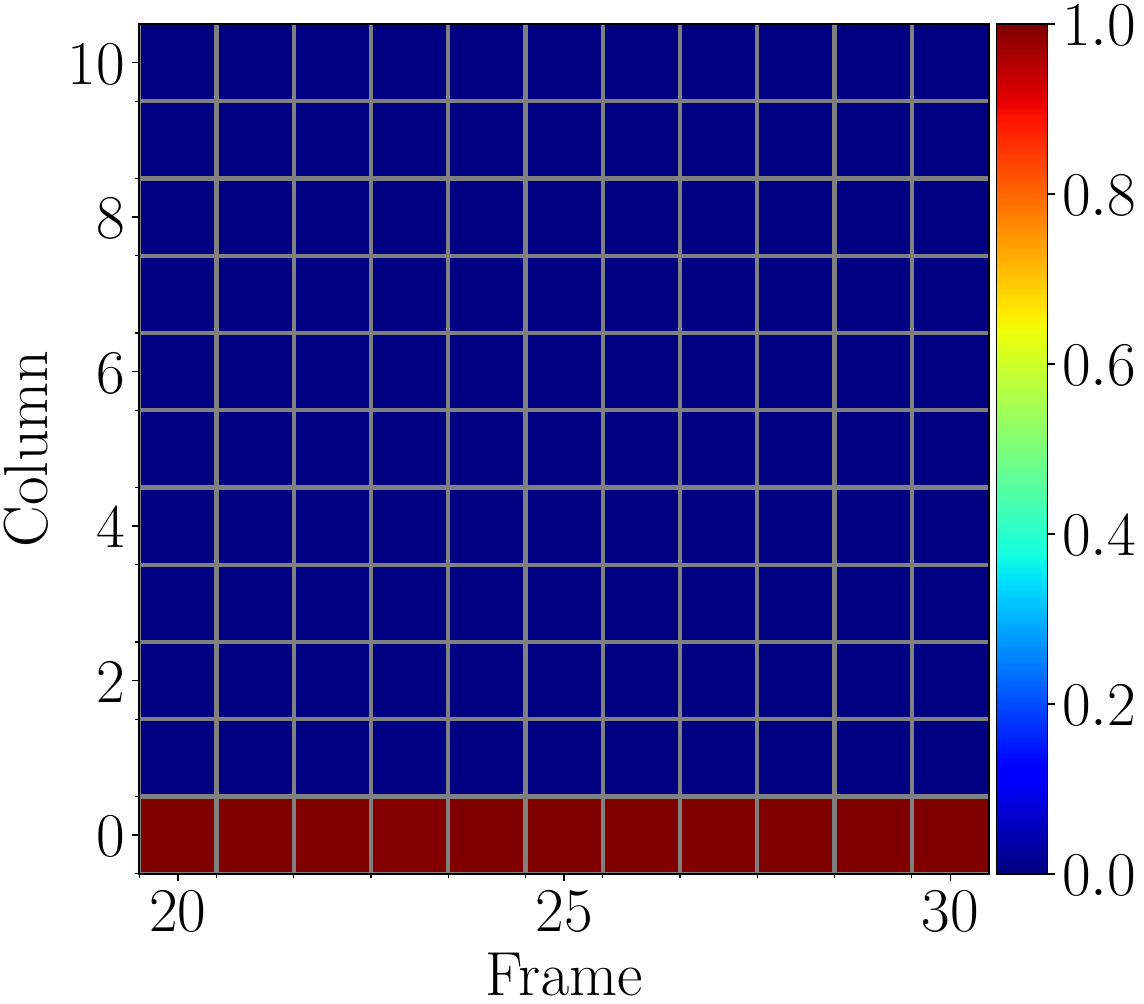}
\end{tabular}
% Red marker (nanoparticle boundary)    \hspace{4em}  vacuum pixel (green)\\
% \vspace*{-0.9em}
% %% 0.461 = 0.3/0.65
% \subfloat[]{\includegraphics[width=0.48\linewidth]{plots_Dec22/Prob_evolution_particle_2.pdf}}\hfill
% \subfloat[]{\includegraphics[width=0.48\linewidth]{plots_Dec22/Prob_evolution_vacuum_2.pdf}}%
% \end{minipage}
\caption{\textbf{Tracking structural dynamics.} Applying the proposed framework at a fixed location over a period of time allows us to monitor the dynamic changes in the atomic structure of a nanoparticle. The graphs show the evolution of the probability assigned to each possible depth by the model at the two locations indicated by the red and green markers in the top image of Figure~\ref{fig:grad_real}. On the left, the estimate at the red marker, which is at the boundary of the nanoparticle, oscillates between depths 0 and 1, identifying dynamic instability in the atomic structure. In contrast, on the right, the depth estimate at the green marker,
which is in the vacuum, is constantly equal to 0.}
\label{fig:dynamics}
\end{figure}
% \begin{figure}[htbp]
%     \centering
%   \subfloat[Input]{%
%        \includegraphics[width=0.5\linewidth]{plots_Dec22/real_particle_output_with arrows.pdf}}
% \\\vspace{-1em}
%   \subfloat[Particle pixel]{%
%         \includegraphics[width=0.49\linewidth]{plots_Dec22/Prob_evolution_particle.pdf}}\hfill
% %\\\vspace{-1em}
%   \subfloat[Vacuum pixel]{%
%         \includegraphics[width=0.49\linewidth]{plots_Dec22/Prob_evolution_vacuum.pdf}}
%         \caption{\textbf{Results on real data: }Applying the segmentation network, frame by frame, to the acquired TEM video allows one to investigate the dynamics of the particle. $(a)$ The atomic column value of the pixel located on the particle oscillates for an initial period, before stabilizing on 1 for a longer period of time. $(b)$ The atomic column value of the pixel located on the vacuum is much more stable.}
% \label{fig:output_real}
% \end{figure}
% \subsubsection{Gradient Analysis }
% \label{sec:analysis}
% Similar to \ref{sec:sim_grad}, we investigated the behavior of the gradient of the neural network with respect to the noisy input. We observe similar qualitative behavior where the gradient on pixels belonging to the nanoparticles have a larger support, while vacuum particles are more localized.

Identifying changes in the atomic structure of nanoparticles is crucial to understand their functionality. An important application of the proposed framework is to monitor such structural dynamics. Figure~\ref{fig:dynamics} provides a proof of concept. The plots show the temporal evolution of the estimated depth probabilities at the two points marked in red and green in the top image of Figure~\ref{fig:grad_real}. On the nanoparticle boundary, the estimated depth alternates between 0 and 1, indicating dynamic structural instability. At the green marker (in vacuum), the estimate is completely stable and equal to 0.

\section{Conclusion and Discussion}
\label{sec:summary}

In this work we introduce a method to recover 3D atomic-scale information from noisy transmission electron microscopy (TEM) images using deep learning. The key idea is to cast depth estimation as a semantic segmentation task and train a deep convolutional neural network on simulated, noise-corrupted data to produce pixel-wise depth maps. Our results on both simulated and real TEM images suggest that this is a promising approach to obtain 3D structural information.

%In this work, we introduce an innovative methodology for extracting details at the three-dimensional atomic level, especially the depth of the atomic columns in CeO2 nanoparticles, from transmission electron microscopy (TEM) images that are significantly affected by Poisson noise. 
%To accomplish this, we developed a deep neural network through simulation-driven training, aimed at interpreting TEM images of nanoparticles and estimating column depth. 

%The network performance was evaluated on both simulated and experimental datasets, yielding initial promising outcomes with real-world data. Upon analysis, we discovered that the network efficiently utilizes extensive-range information to accurately reconstruct the local 3D structure. Our results indicate that the network has beneficial characteristics, such as appropriate calibration and resilience to noise levels encountered during training. This approach shows great potential for enhancing the understanding of electrochemical processes, including the mechanisms of transport and exchange within active nanoparticles. 

Future work will include the creation of more sophisticated datasets that encompass diverse imaging configurations, including tilt angles, rotational variations, and scale enhancement. Similarly, it is imperative to expand the simulation datasets to include atomic defects and various elemental compositions. In addition, there is a need for further evaluation on real experimental data. Finally, from the methodological point of view, an important challenge is how to use real experimental data to train and evaluate the underlying deep-learning models, as in unsupervised~\cite{udvd,crozier2025visualizing,morales2023evaluating}. A promising direction is to combine an unsupervised denoising loss with a supervised segmentation loss as in \cite{denoiseg}.

\subsection*{Acknowledgements}
The authors acknowledge ASU Research Computing and NYU HPC for providing high performance computing resources, and the electron microscopes in the Eyring Materials Center at Arizona State University. We thank Joshua Vincent for providing the experimental data used in Section~\ref{sec:real_dataset}. We gratefully acknowledge financial support from the National Science Foundation. NSF OAC 1940263 and 2104105 supported PAC and RM, NSF DMR 1840841 supported MT, NSF OAC 2103936 supported ML and CFG.

\bibliographystyle{IEEEtran}

\bibliography{EM3D_bib}

\appendix

\section{Network Architecture}
 \label{sec:app_network_architecture}

The model is based on a U-Net convolutional encoder--decoder architecture with skip connections. The encoder path consists of an initial double-convolution block followed by a sequence of downsampling stages, each implemented as $2\times2$ max pooling followed by two $3\times3$ convolutional layers with replication padding, batch normalization, and ReLU activations. The number of resolution scales is chosen at 6, allowing the network depth and receptive field to be adjusted. At each scale, feature maps are either maintained or progressively increased in channel dimension. The number of base channels controls the fundamental width of the network and serves as the primary scaling factor for channel dimensionality throughout the architecture. The decoder mirrors the encoder, using bilinear upsampling (or optionally transposed convolutions) followed by concatenation with the corresponding encoder features and additional double-convolution blocks. This design enables the network to combine coarse contextual information with fine spatial details through multi-resolution skip connections.

After the standard U-Net decoding path, the architecture optionally includes two additional median downsampling convolutional blocks that further process the reconstructed full-resolution features before the final output layer. These blocks reduce spatial resolution by a factor of four and increase channel dimensionality prior to a final $1\times1$ convolution that maps features to the desired number of output channels, corresponding to the number of possible atomic column depth values.
\end{document}